\documentclass{article}

\usepackage{arxiv}
\usepackage{cite}
\usepackage{amsmath,amssymb,amsfonts}
\usepackage{algorithmic}
\usepackage{graphicx}
\usepackage{textcomp}
\usepackage[acronym]{glossaries}
\usepackage[capitalise,noabbrev]{cleveref}

\newcommand{\norm}[1]{\left\lVert#1\right\rVert}
\newacronym{ann}{ANN}{Artificial Neural Network}
\newacronym{snn}{SNN}{Spiking Neural Network}
\newacronym{lif}{LIF}{Leaky-Integrate and Fire}
\newacronym{alif}{ALIF}{Adaptive Leaky-Integrate and Fire}
\newacronym{bptt}{BPTT}{Back-Propagation Through Time}
\newacronym{odesa}{ODESA}{Optimized Deep Event-driven Spiking neural network Architecture}
\newacronym{wta}{WTA}{Winner-Take-All}
\newacronym{srm}{SRM}{Spike Response Model}
\newacronym{stdp}{STDP}{Spike-Timing-Dependent Plasticity}
\newacronym{iou}{IOU}{Intersection Over Union}

\title{An optimised deep spiking neural network architecture without gradients}

\author{
  Yeshwanth ~Bethi \\ 

  ICNS\\
  Western Sydney University\\
  \texttt{Y.Bethi@westernsydney.edu.au} \\

   \And
   
  Ying ~Xu \\ 

  ICNS\\
  Western Sydney University\\
  \texttt{Ying.xu@westernsydney.edu.au} \\

   \And
   
  Gregory ~Cohen \\ 

  ICNS\\
  Western Sydney University\\
  \texttt{G.Cohen@westernsydney.edu.au} \\

 \And
    
  Andre van ~Schaik \\ 
  ICNS\\
  Western Sydney University\\
  \texttt{A.vanschaik@westernsydney.edu.au} \\

    \And
  
  Saeed ~Afshar \\ 
  ICNS\\
  Western Sydney University\\
  \texttt{S.Afshar@westernsydney.edu.au} \\

}
\begin{document}
\maketitle

\begin{abstract}
We present an end-to-end trainable modular event-driven neural architecture that uses local synaptic and threshold adaptation rules to perform transformations between arbitrary spatio-temporal spike patterns. The architecture represents a highly abstracted model of existing \acrfull{snn} architectures. The proposed \acrfull{odesa} can simultaneously learn hierarchical spatio-temporal features at multiple arbitrary time scales. \acrshort{odesa} performs online learning without the use of error back-propagation or the calculation of gradients. Through the use of simple local adaptive selection thresholds at each node, the network rapidly learns to appropriately allocate its neuronal resources at each layer for any given problem without using an error measure. These adaptive selection thresholds are the central feature of \acrshort{odesa}, ensuring network stability and remarkable robustness to noise as well as to the selection of initial system parameters. Network activations are inherently sparse due to a hard \acrfull{wta} constraint at each layer. We evaluate the architecture on existing spatio-temporal datasets, including the spike-encoded IRIS, latency-coded MNIST, Oxford Spike pattern and TIDIGITS datasets, as well as a novel set of tasks based on International Morse Code that we created. These tests demonstrate the hierarchical spatio-temporal learning capabilities of \acrshort{odesa}. Through these tests, we demonstrate \acrshort{odesa} can optimally solve practical and highly challenging hierarchical spatio-temporal learning tasks with the minimum possible number of computing nodes.
\end{abstract}


\section{Introduction}
\label{sec:introduction}
Over the last decade, Deep Learning using \acrfull{ann}s
has been adapted to almost every computational field. The vast adoption of Deep Learning can be attributed to the simple error back-propagation algorithm \cite{rumelhart1986learning} that enabled auto-differentiation for the ANN models. The error back-propagation algorithm solves the credit-assignment problem for any arbitrary cascade of layers, activation functions, and any given loss function, as long as they are all differentiable. The power of such modularity gave rise to complex hierarchical models which could adapt to any domain of data such as numerical, images, audio, video, robotics, and natural language. Modular ANNs are also capable of handling multi-modal data from different domains. Yet, the error back-propagation rule which underpins ANNs does not fit well with our understanding of the form and function of biological \acrshort{snn}s, which were the original inspiration of ANNs. This troubling incongruity is a major focus of research in computational neuroscience and has motivated many attempts at adaptation, reconciliation, and re-interpretation of the computations involved in both artificial and biological neural networks \cite{Stork1989}.

There are several reasons why the error back-propagation algorithm can not plausibly be implemented in biological neural networks \cite{crick1989recent} \cite{grossberg1987competitive}. First, error back-propagation requires a numerically precise error measurement for a batch of data. To date, no evidence for such batch processing or numerically precise error measurements has been uncovered in the brain and given our mature understanding of the behaviour of biological neural networks, such evidence is unlikely to emerge. Second, successful error back-propagation requires global, precise, and repeated propagation of error measures through all the involved computational nodes in a network - beginning from the inputs to the highest layers and back. This has been dubbed the 'weight transport problem', as the weights of higher layers have to be made available to the lower layers for successful backpropagation of error values \cite{grossberg1987competitive}. Again, no evidence for such processes has been, or is likely to be, found. The third and most crucial pillar of error back-propagation is the differentiability requirement for all the constituent components of a given network. Indeed, it is this very aspect of the error back-propagation algorithm which makes it difficult to use on non-differentiable data domains like spatio-temporal spikes patterns which the brains use as the primary mode of computation and communication. This leads to the biggest problem in the use of error back-propagation in computational neuroscience namely, the credit-assignment problem. 

 There have been multiple attempts at approximating error back-propagation and applying gradient descent to \acrshort{snn} architectures. SpikeProp \cite{Bohte2000} was among the first works to derive a supervised learning rule for \acrshort{snn}s from the error back-propagation algorithm. Tempotron \cite{gutig2006tempotron} was introduced in which the error back-propagation was applied by defining loss functions based on the maximum voltage and the threshold voltage of the output neurons. Chronotron \cite{florian2012chronotron} was introduced as an improvement over Tempotron by using a new distance metric between the predicted and target spike trains. More recent works applied error back-propagation to \acrshort{snn} architectures by using different surrogate gradients for the hard thresholding activation functions of the spiking neurons \cite{Neftci2019}\cite{bellec2018long}\cite{bellec2020solution}\cite{zenke2018superspike}. However, they don't address how biology can realize the computation of gradients and their access to the neurons involved in the computation. Furthermore, we don't have evidence on how batching of data happens in biology, and most of the gradient descent approaches rely on batching the data. Despite the lack of bio-plausibility, the error back-propagation based approaches have been adopted in computational neuroscience as useful alternative tools to discover the required connectivity in \acrshort{snn}s for a given specific task \cite{zipser1988back} \cite{mcclelland1995there} \cite{zenke2021remarkable}.

Feedback alignment has been used as an alternative to error back-propagation for \acrshort{snn}s in \cite{lillicrap2016random} to solve the 'weight transport' problem. Feedback alignment shows that multiplying errors by random synaptic weights is enough for effective error back-propagation without requiring a precise symmetric backward connectivity pattern. There have been parallel investigations of more bio-plausible local learning rules for \acrshort{snn}s which do not require access to the weights of other neurons in the network. This set of learning rules for \acrshort{snn}s can be characterized as synaptic plasticity rules which use \acrfull{stdp} in some form. \acrshort{stdp} rules have more commonly been used for extracting features from spike trains in an unsupervised manner. There have also been probabilistic \cite{tavanaei2016acquisition} and reinforced \cite{Mozafari2018} variants of \acrshort{stdp} to learn discriminative features which can assist classification. \acrshort{stdp} rules are also commonly used along with \acrshort{wta} rules to promote competition among neurons and reduce the redundancy in the learnt features \cite{vigneron2020critical}. Paredes-Vallés et.al \cite{paredes2019unsupervised} used a homeostasis parameter that controls the excitability of neurons in combination with \acrshort{stdp} and \acrshort{wta} rules to promote stability. Supervised Hebbian Learning (SHL) \cite{legenstein2005can} and ReSuMe \cite{ponulak2010supervised} were one of the first works to use \acrshort{stdp} rules to perform supervised learning in single layer \acrshort{snn} networks. ReSuMe implements a spiking version of Widrow-Hoff rule\cite{widrow1960adaptive} for rate-coded \acrshort{snn}s and using \acrshort{stdp} and anti-\acrshort{stdp} processes. Spike Pattern Association (SPAN)\cite{mohemmed2012span} is another learning method that adopted the Widrow-Hoff rule for spike sequence learning. \cite{sporea2013supervised} extend the ReSuMe to multiple layers and approximate the gradient descent update step of the intermediate layers to an \acrshort{stdp} process. Taherkhani et.al, 2018\cite{taherkhani2018supervised} proposed another modification to the ReSuMe rule to extend it to multi-layer \acrshort{snn}s.

Another set of learning algorithms that gained momentum recently is applying evolutionary methods for \acrshort{snn} optimization. Some works have been solutions for \acrshort{snn} network structure optimization \cite{schliebs2013evolving} \cite{wysoski2006adaptive} \cite{Lobo2018} \cite{Kasabov2019} and others have been for synaptic weights optimization \cite{belatreche2006evolutionary} \cite{Pavlidis2005} \cite{Vazquez2011}. 
Apart from these broader themes, there have been other works that used practical mathematical solutions to the learning problem. For instance, SpikeTemp\cite{wang2015spiketemp} used rank-order based learning for \acrshort{snn}s. Multi-Spike Tempotron (MST)\cite{gutig2016spiking} used a technique called aggregate label learning that can learn predictive cues or features. MST was followed up with a computationally simpler version called Threshold-Driven Plasticity (TDP)\cite{yu2018spike}. Both MST and TDP update weights by calculating the gradients with respect to the threshold of a neuron and finding the optimal threshold for a desired number of spikes from a neuron. Membrane Potential Driven Aggregate Learning (MPD-AL) \cite{zhang2019mpd} was proposed as an alternative version of aggregate learning which used gradients with respect to the membrane potential. The role of neuromodulators in synaptic learning was explored through three-factor learning rules in \cite{farries2007reinforcement}\cite{brea2013matching}\cite{friedrich2011spatio}\cite{florian2007reinforcement}\cite{jimenez2014stochastic}. The key motivation to use three-factor learning rules is to extend the functionality of \acrshort{stdp} beyond unsupervised learning, which by design, neglects any information regarding “reward", “punishment,” or “novelty” during learning \cite{fremaux2016neuromodulated}.

Alongside the regular \acrshort{snn} architectures, with the recent popularity of neuromorphic vision sensors \cite{lichtsteiner2008128}, multiple event-based neural architectures have been proposed \cite{sironi2018hats}\cite{lagorce2016hots}.  
Afshar et.al, 2020 \cite{afshar2020event} proposed Feature Extraction using Adaptive Selection Thresholds (FEAST) as an unsupervised feature extraction algorithm for event-based data using exponential time surfaces and spiking neuron-like units which have individual selection thresholds. In the FEAST algorithm, each feature unit is represented by a weight vector for all its inputs and a selection threshold that facilitates equal activation of the features during online learning. In this paper, we propose to use these adaptive selection thresholds for multi-layer supervised learning and propose our method as a simple solution to the credit assignment problem. Our proposed method is the first to not require the transport of weights across neurons in the network or random connections for error propagation. We achieve all feedback to earlier layers using precisely timed binary attention signals which signal "reward" and "punishment" of the recently active neurons.

\section{Background and Related Work}

\subsection{Time Surfaces}\label{subsec:Timesurf} Tapson et al. 2013
\cite{tapson2013synthesis} proposed the use of exponentially decaying kernels for processing event data produced by neuromorphic vision sensors. We shall use the terminology introduced by Lagorce et al. 2016 \cite{lagorce2016hots} and refer to these kernels as time surfaces. The time surface is the trace of events per each input channel which is updated only when an event arrives at a channel. An event from an event-based sensor can be described as: 
\begin{equation}\label{eqn:EventEqn}
e_i=(x_i,y_i,t_i,p_i)
\end{equation}

\cref{eqn:EventEqn} describes an event from a pixel location $(x_i,y_i)$ as the coordinates on the sensor, with polarity $p_i$, arriving at time $t_i$. A similar notation can be used to represent a spike as an event: 

 \begin{equation}\label{eqn:SpikeEqn}
s_i = (c_i,t_i)
\end{equation}

\cref{eqn:SpikeEqn} represents a spike $s_i$ from channel $c_i$ at time $t_i$. We can use the time surfaces to keep the trace of spikes from a spiking source. 

The exponential time surface $S_{t}[i]$ of a channel $i$ at time $t$ with a time constant of $\tau$ can be calculated as follows:

\begin{equation}\label{eqn:TimeSurface}
        S_{t}[i] = P[i]*e^{\frac{-(t -TS[i])}{\tau}}
\end{equation}

Where $P[i]$ in \cref{eqn:TimeSurface} is the last updated potential of the channel $i$ and $TS[i]$ holds the timestamp of the last spike to have occurred at the channel $i$. If a new spike $(i,t_i)$ occurs at the channel $i$ then the potential of the channel $P[i]$ is updated according to \cref{eqn:Polarity,eqn:Timestamp}

\begin{align}
    \underset{new}{P[i]} &= \underset{old}{P[i]}*e^{\frac{-(t_i -TS[i])}{\tau}} + c \label{eqn:Polarity}\\
    \underset{new}{TS[i]} &= t_i \label{eqn:Timestamp}
\end{align}

Where $c$ is the constant by which the potential is increased for each new spike at a channel. $c$ is generally set to $1$.

This formulation of time surfaces for spikes is simply a reformulation of the Excitatory Post Synaptic Potential due to an input spike to a biological neuron as described in the Steins model\cite{Stein1967}\cite{Stein1965}, which jumps by an amount $w_k$ on the arrival of a spike and decays exponentially thereafter. 

\begin{align}
    \Delta u(t) &= w_k*\epsilon(t-t_{k}^{f}) \\
    \epsilon(t) &= e^{\frac{-t}{\tau_m}}
\end{align}

A neuron $n$ can then be parameterized by a weight vector $W_n$ representing the synaptic weights to all the input channels $N_c$ and a spiking threshold $\theta_n$. The total contribution to the membrane value $v_n$ of the neuron $n$ by all input spikes arrived up until a time can be formulated as the dot product of the normalized time surface of all channels and the normalized weight vector. The normalization of both the potential and the weights ensures the membrane value $v_n$ is between 0 and 1. The membrane value $v_n$ is calculated according to \cref{eqn:context,eqn:membraneval}. $C$ in \cref{eqn:context} represents the time surface context which is the normalized time surface of all input channels. The time surface context $C$ represents the recent activity from all the input channels.
\begin{align}
    C &= \frac{S}{\norm{S}} \label{eqn:context}\\
    v_{n} &= \underset{(1 \times N_c)}{W_{n}}\cdot \underset{(1 \times N_c)}{C} \label{eqn:membraneval}
\end{align}

Afshar et al. 2020  \cite{afshar2020event} introduced an algorithm to extract features from event-data using layers of such neuronal units in an unsupervised manner. In addition to the weights which represent the features, each neuron has a threshold as an additional parameter. For every input event, the dot product ($v_n$) of the time surface context ($C$) and the weight ($W_n$) of a neuron is calculated. The dot products of all neurons are then compared to their respective thresholds. Out of the neurons with the dot products greater than their respective thresholds($v_n \geq \theta_n$), the neuron with the largest dot product is considered the winner for the given input event. If none of the dot products crosses their respective neuron thresholds, the thresholds of all the neurons are reduced by a pre-defined fixed value. On the contrary, if a matching neuron is found, the feature/weight vector $W_n$ is updated with the current event context $C$ using an exponential moving average, and the threshold of the neuron is increased by a fixed value. The thresholds and weights of other neurons are left unchanged.

The adaptation of the selection thresholds promotes homeostasis and facilitates equal activation of the feature neurons in response to the data. This Feature Extraction using Adaptive Selection Thresholds (FEAST) is an online learning method that clusters the incoming event contexts of all the events into $N$ clusters where $N$ is the number of neurons used in a FEAST layer.

Lowering the threshold of a feature neuron makes that neuron more receptive to new event contexts, whereas increasing the threshold makes a feature neuron more selective. FEAST treats each incoming event with equal priority as there is no information regarding the importance of individual events. This results in the features representing the most commonly observed spatio-temporal patterns in the input data. However, learning the most commonly occurring features may not be ideal for tasks that depend on rarer task-specific features.

\subsection{Abstraction of Spiking Neural Networks}
The space of possible \acrshort{snn} architectures can be characterized by the different models of neurons, synapses, learning rules, and network architectures used in them. In this space, there is often a trade-off between the biological plausibility and practical applicability of the models. Network models attempting to demonstrate biological plausibility through detailed phenomenological modelling from the voltage-gated ion channels to the delays at the neuronal synapses tend to be limited in their performance and utility in the context of challenging machine learning tasks. The computational cost of these models increases with the bio-plausibility of the model. Different neuronal models have been proposed which approximate and abstract the details of these complexities with easy-to-handle mathematical and probabilistic models \cite{fourcaud2003spike} \cite{gerstner2002spiking} \cite{izhikevich2004model} \cite{kasabov2010spike}. \acrfull{lif} neuron model \cite{Stein1965} and specifically the \acrfull{srm} \cite{jolivet2004generalized} are among the most popular choices of neuron models in the \acrshort{snn}s, even though the degree to which they explain the neuronal dynamics is limited compared to other models like Hodgkin-Huxley \cite{hodgkin1952quantitative} or Izhkevich \cite{izhikevich2004model} models. Their vast adoption can be attributed to their analytical tractability and computational simplicity compared to other neuronal models.
But even the \acrshort{snn} models which use simpler neuronal models like \acrshort{lif} or \acrfull{alif} neurons \cite{jolivet2006predicting} require additional complexities such as Excitatory-Inhibitory Balance, and the right amount of lateral excitation and inhibition to instil behaviours like \acrshort{wta}. These complex processes make it difficult to scale up the simulations of the multi-layered \acrshort{snn}s and limit the exploration of broader system-level learning mechanisms of the \acrshort{snn}s as there are a lot of variables in the system.
In the same way, that time surfaces represent simplified hardware friendly abstractions of the EPSP, the FEAST network can be best understood as a highly abstracted, functionally equivalent, modular implementation of a well-balanced excitatory \acrshort{snn} with inhibitory feedback leading to a winner take all operation at a single layer. In this way, a FEAST layer represents a neuron group. Picking only one winner in each layer of FEAST for any input event is a proxy for hard \acrshort{wta} motif in a neuron group, without requiring any forms of inhibition. Simpler and computationally easier abstract \acrshort{snn} models like FEAST can help us explore more system-level learning rules in \acrshort{snn}s without having to worry about problems like achieving EI balance and promoting or removing oscillations in the networks. Just like Address-Event Representations (AER) being used in Neuromorphic hardware to facilitate the communication in \acrshort{snn}s, we can use novel abstractions like FEAST to explore the space of local learning rules in Spiking Neural Architectures. Continuing in this approach and extending it, the \acrfull{odesa} introduced in this paper, represents a method to locally train hierarchies of well-balanced EI networks on event-based data in a supervised manner. In this way, the abstracted \acrshort{snn} which \acrshort{odesa} represents can be used to rapidly explore a wide range of multi-layered \acrshort{snn} models for real-world online supervised learning applications. 

\section{Materials and Methods}
The aim of \acrshort{odesa} is to use a multi-layered spiking neural architecture and train it to map any input spatio-temporal spike pattern $X_{train}$ to any output spatio-temporal spike pattern $Y_{train}$ and to do so entirely using binary signals, and without having access to the weights of other neurons or batching of input data. The latter restriction not only makes \acrshort{odesa} a useful framework for studying local learning biological \acrshort{snn}s but also allows local online training in neuromorphic hardware implementations of such networks. The \acrshort{odesa} architecture can contain multiple hidden layers with different time constants to learn hierarchical spatio-temporal features simultaneously at different timescales to support an output layer consisting of classification readout neurons which generate the desired spike in $Y_{train}$. 

\subsection{Classification using adaptive selection thresholds} \label{subsec:ClassificationSection}
The output classification layer in \acrshort{odesa} has $k*N_c$ neurons ($k=1,2,3,\dots$) for a classification task with $N_c$ classes. The output layer is divided into $N_c$ groups, each responsible for spiking for their respective classes. For any given input spike to the layer, only one neuron (out of $k*N_c$ neurons) can spike.

The threshold adaptation in \acrshort{odesa}'s output layer is driven by the supervisory spike signal $Y_{train}$ for a given input spike stream $X_{train}$. Considering that \acrshort{odesa} is event-driven, it is assumed that there exists an input spike $i_t$ in $X_{train}$ at time $t$ for every output spike $o_t$ in $Y_{train}$. The labelled input spike $i_t$ in $X_{train}$ which has an output label spike $o_t$ associated with it, is treated with additional attention. For the labelled input spike $i_t$, if there is no spike from the respective class neuron group responsible for the current class of the supervisory spike $o_t$, the thresholds for all the neurons in the class group are lowered. If there is a spike from any of the neurons in the class group, the winner neuron's weights are updated with the input spike's event context, and its threshold is updated. Alternatively, in the absence of a spike from the correct class group, the thresholds of all the neurons in the group are reduced. This weight update and threshold increase in a neuron can be thought of as 'rewarding a neuron' for its correct classification. A decrease in the threshold of a neuron to make it more receptive can be viewed as 'punishing a neuron' for not being active. 

The threshold increase step in \acrshort{odesa} is different from that proposed in Afshar et al. 2020 \cite{afshar2020event}. Rather than the fixed value used in the previous work, it is an adaptive value that depends on the dot product of the context with the weight according to \cref{eqn:ThreshDiff,eqn:OriginalThreshUpdate}. The $\Delta \theta_{n}$ in \cref{eqn:ThreshDiff} is never negative because for any winner neuron $n$, $v_{n}$ is always at least as high as $\theta_{n}$ to win. This new threshold adaptation ensures that threshold ($\theta_{n}$) of a neuron moves asymptotically towards, but never reaches 1 as the model gets better at classification. This modification speeds up the threshold adaptation operation, while simultaneously improving the stability of the system by having a proportional increase in the threshold based on the membrane value ($v_{n}$) at the time of winning an input spike. \crefrange{eqn:contextFormation}{eqn:OriginalThreshUpdate} show the weight and threshold adaptation of a winner neuron $n$ if it belongs to the correct class group. 

\begin{align}
    C &= \frac{S(i_t)}{\norm{S(i_t)}} \label{eqn:contextFormation} \\ 
    v_{n} &= W_{n}\cdot C \\
    \Delta W_{n} &= C - W_{n}  \\
    W_{n} &= W_{n} + \eta * \Delta W_{n} \label{eqn:OriginalWeightUpdate}\\
    W_{n} &= \frac{W_{n}}{\norm{W_{n}}}  \\
    \Delta \theta_{n} &= v_{n} - \theta_{n}  \label{eqn:ThreshDiff}\\ 
    \theta_{n} &= \theta_{n} + \eta_{thresh}*\Delta \theta_{n} \label{eqn:OriginalThreshUpdate}
\end{align}
\cref{eqn:punishGroup} shows the punishment of neuron group which has not spiked by lowering the thresholds of the group. $\Theta_{label}$ represents the thresholds of all the $k$ neurons in the class group that corresponds to the label of the input spike. $\theta_{drop}$ is the fixed value by which the thresholds are reduced.

\begin{equation}
    \Theta_{label} = \Theta_{label} - \theta_{drop} \label{eqn:punishGroup}
\end{equation}

The \acrshort{wta} constraint in the \acrshort{odesa} layers ensures that there can be at most one winner neuron for each input spike to a layer. This creates competition between the neurons to capture regions of the input space that precede output labels. In this way, neurons in each layer attempt to only learn the spatio-temporal features that are crucial in discriminating one class from another class ensuring that the neuron groups don't learn features that are common between two different classes. 

The rewarding and punishing of neurons based on their activity with respect to the label spike stream is the key element of learning in \acrshort{odesa}. 

\subsection{Multi-Layer Supervision through Spike-Timing-Dependent Threshold Adaptation}

\begin{figure*}[h]
\centering
\includegraphics[width=0.85\textwidth]{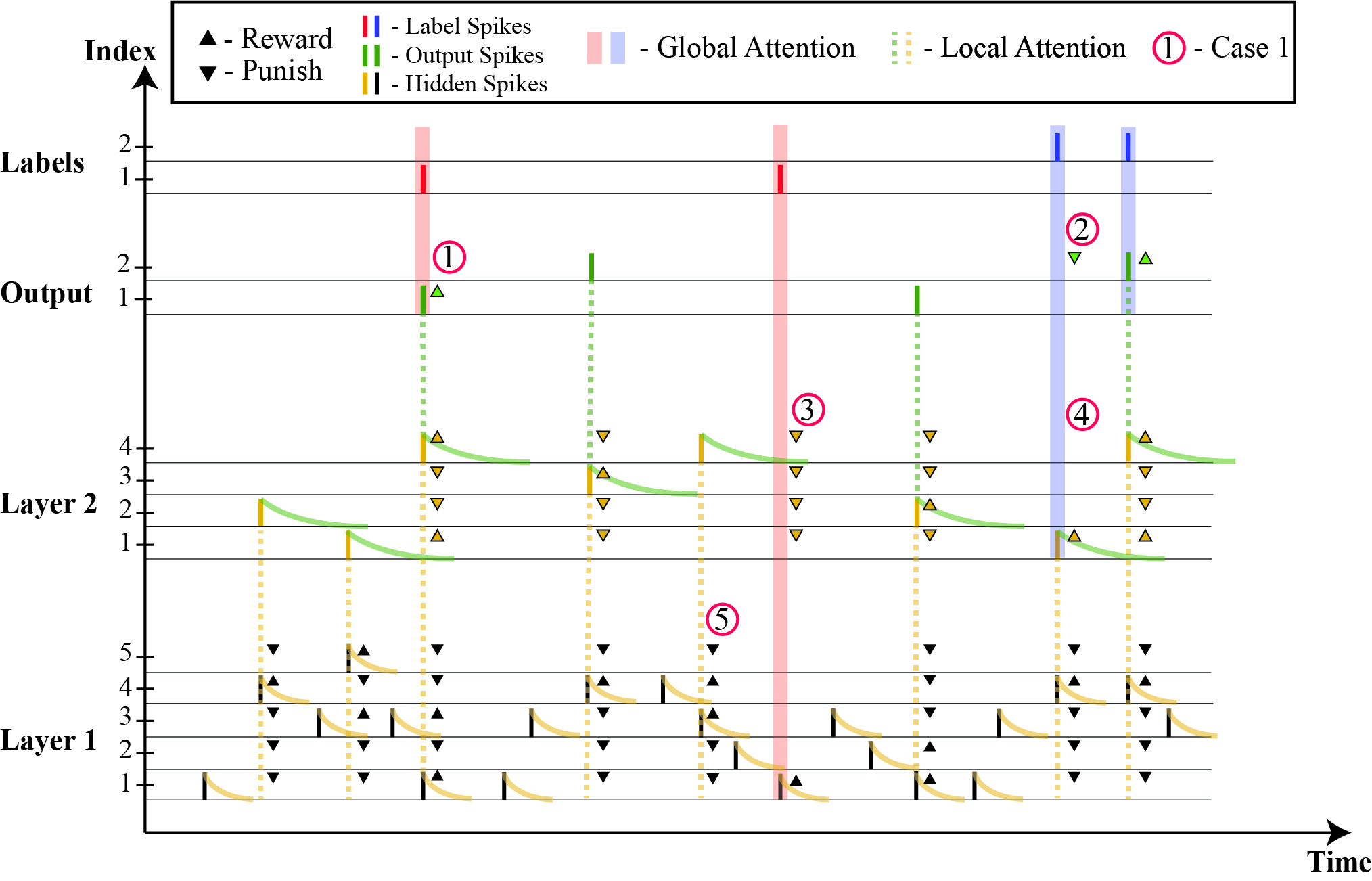}
\caption{Multi-Layer Supervision in \acrshort{odesa} using Spike-Timing-Dependent Threshold Adaptation. The shaded vertical lines represent the binary Global Attention Signal generated for each output label spike, and the dotted vertical lines represent the binary Local Attention Signals sent to each layer from its next layer. The up and down arrows represent the reward and punishment of the individual neurons. Case 1: The predicted output spike matches the label spike, and the corresponding output neuron is rewarded. Case 2: The corresponding output neuron for the correct class is punished as it failed to spike in the presence of input from Layer 2. Case 3: All neurons in Layer 2 are punished as they failed to spike for an input spike from Layer 1 in presence of the Global Attention Signal. Case 4: The active neuron in Layer 2 is rewarded in presence of the Global Attention Signal. Case 5: The neurons with trace above the recency threshold are rewarded and the other neurons are punished in the presence of Local Attention Signal from Layer 2.}
\label{fig:ODESASTDTP}

\end{figure*}

Decaying event kernels, time surfaces, and the EPSPs they represent are all imperfect as memory units since they can map a wide range of spike trains onto the same analog value at a channel, thus losing potentially critical information, especially when there are multiple spikes per input channel. This poses a serious problem for real-world tasks which depend on features that occur at multiple timescales and which generally contain an arbitrary number of information-carrying spikes per channel. Furthermore, in most real-world tasks, a shared collection of low level features, when combined in varying ways in time and feature-space, maps the input data to the desired output. Thus, hierarchical layers are often required to solve complex tasks that require associations at different feature levels and timescales.

Multiple spiking layers, however, pose a new problem to learning not present in simple one layer networks such as FEAST. This is a problem of credit assignment. Because the thresholding operation in \acrshort{snn}s is non-differentiable, other works have used various versions of gradient approximation and \acrfull{bptt} \cite{zenke2018superspike}\cite{Neftci2019}\cite{bellec2018long}\cite{bellec2020solution}. In contrast, \acrshort{odesa} solves this problem without the use of gradients, by using the activity of the next layer as the supervisory signal for the current layer.

Each neuron in the hidden layer of an \acrshort{odesa} network has a trace of its latest activation. \cref{eqn:TraceEq} describes the trace $\Lambda^{l}_{t}[n]$ of a neuron $n$ in layer $l$ at time $t$. Just like time surfaces, updates of the trace of a layer ($\Lambda^{l}$) are event-driven. $t_{n}$ is the time of neuron $n$'s most recent spike. The trace $\Lambda^{l}_{t}[n]$ acts as a measure that indicates the neuron $n$'s recent activity at any time $t$. The trace of a layer $\Lambda^{l}_{t}$ is used to find the neurons that participated in generating a spike in the next layer and reward or punish them accordingly. The time constant of the trace $\Lambda^{l}$ is equal to the time constant of the neurons in its next layer $\tau_{l+1}$, whereas the time constant $\tau_l$ for neurons in a layer $l$ is used to decay the inputs to layer $l$ via time surface $S^l$. Thus in general, the time constant of layer $l$'s trace $\tau_{l+1}$ is not equal to that layer's time surface time constant $\tau_{l}$. In our experiments, we have used the same time constant for all the neurons in a layer.  When any neuron $n_{l+1}$ in the next layer spikes (post-synaptic spike) for an input spike (pre-synaptic spike) from the current layer ${l}$, a local binary attention signal $A^{l+1}[i]$ is passed to the current layer ${l}$ from the next layer ${l+1}$ indicating activity in the next layer. The current layer ${l}$ uses the local attention signal to reward its recently active neurons (whose trace $\Lambda^{l}_{t}[n] \geq \Phi$) and punishes its inactive neurons ($\Lambda^{l}_{t}[n] < \Phi$) where $\Phi$ is the trace recency threshold. In this work, an arbitrarily chosen trace recency threshold value of $\Phi=0.1$ is used throughout. The reward and punishment of the neurons in this layer are the same as described in \cref{subsec:ClassificationSection}.

\begin{equation}
       \Lambda^{l}_{t}[n] = e^{(\frac{-(t -t_{n}^{f})}{\tau_{l+1}})}\label{eqn:TraceEq}
\end{equation}

This local attention signal driven reward mechanism forces the neurons in each layer to learn features that best support the activity in the next layer. The neurons in the next layer in turn compete to support the activity of neurons in the following layer and so on. The last hidden layer of the network is rewarded by the output layer, and the output layer is rewarded by label spikes which are events that carry the ground truth labels. The supervision of the output layer can be considered the same as the supervision applied to the hidden layers but with a trace that decays instantaneously, i.e., $\tau_{o+1} \rightarrow 0$. The instantly decaying trace ensures the rewarding of the output layer only when it generates a spike precisely at the time of the actual label spike. The post-synaptic spike-timing-dependent threshold adaptation described is the key element to the learning in multi-layered \acrshort{odesa}. This threshold adaptation mechanism is an additional dimension to the learning process apart from the usual synaptic weight adaptation used in \acrshort{stdp}-based \acrshort{snn}s. It helps in regulating the spike activity and utilises all the neuron resources available by promoting equal activation of all the hidden neurons involved in generating an expected output spike. Thus providing the required behavioural complexity needed to solve the credit assignment problem over multiple layers.

Since the \acrshort{odesa} network is event-driven, if a hidden layer $l$ fails to spike for an input spike there will be no spike generated at the output layer that can be used for training. Therefore, in addition to the local attention signal $A^{l+1}[i]$, a Global Attention Signal $G[i]$ is also generated for every labelled input spike. Every layer is expected to spike for a labelled input spike such that an output spike can be generated. Hence, all the neurons in the silent layer $l$ that failed to spike in presence of the Global Attention Signal and an input spike from its previous layer are punished. This rule isolates the layer where the failure to transmit spikes took place and punishes the neurons in that layer by making them more receptive. Additionally, the neurons in every layer which were active during the global attention signal are rewarded. In this way, every layer in a well trained network will generate a spike whenever an output spike is expected. 
\cref{fig:ODESASTDTP} shows the spike-timing-dependent threshold adaptation for a sample network activity. The exponential kernels show the trace of the spikes at each layer and the dotted lines show the local attention signals a layer passes to the layer below it when a spike is generated by one of its neurons. The upward arrows indicate the neurons that are rewarded and the downward arrows indicate the neurons that are punished. The red and blue shaded zones indicate the global attention signals generated when a labelled spike arrives.

\subsection{Overview of learning rules in ODESA}

\begin{figure}[htbp]
\centering
\includegraphics[width=\linewidth]{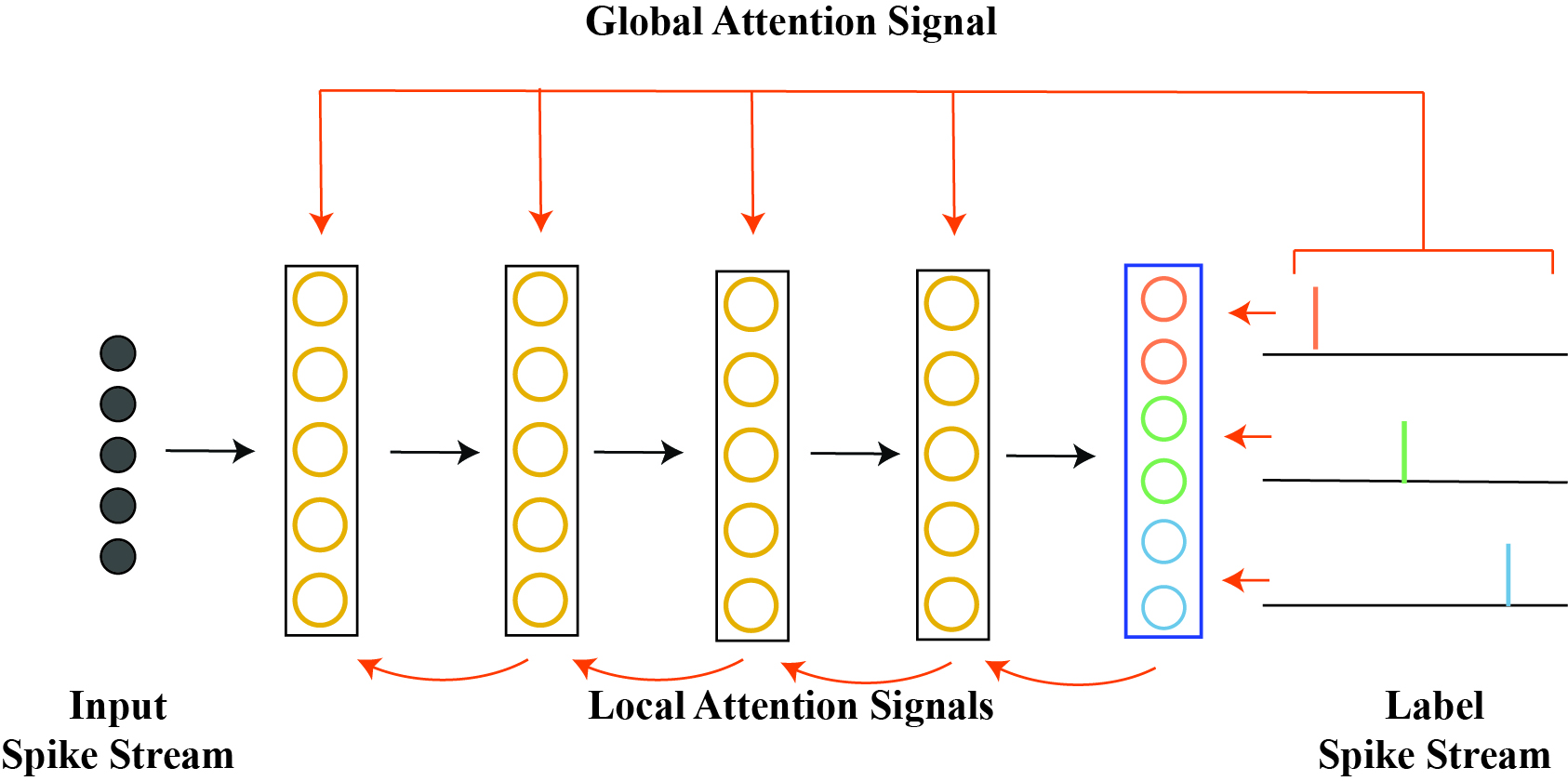}
\caption{Overview of the supervision in multi-layered \acrshort{odesa}}
\label{fig:ODESAOverview}
\end{figure}

\cref{fig:ODESAOverview} shows the operation of a multi-layered \acrshort{odesa} network. The Global Attention Signal $G[i]$ is a binary event that indicates the presence of a labelled spike. The first layer $l$ in the network that stays inactive (Case 3 in \cref{fig:ODESASTDTP}) while having an input spike from the layer below it $l-1$ is punished in the presence of the Global Attention Signal. The Local Attention Signals $A^{1:L}[i]$ are also binary events that indicate the activation of the next layer to the current layer. The recently active neurons ($\Lambda^{l}_{t}[n] \geq \Phi$) in the previous layer are rewarded (weight update and threshold adaptation as in \crefrange{eqn:OriginalWeightUpdate}{eqn:OriginalThreshUpdate}) and the inactive ones are punished (threshold decreased as in \cref{eqn:punishGroup}). This supervision is different to traditional back-propagation techniques which require access to all the higher layers' activity to find the gradients for a given layer. \acrshort{odesa} training continues even when one of the later layers in the network goes silent. This spike-timing-dependent threshold adaptation equips \acrshort{odesa} with the ability to learn features simultaneously at all layers irrespective of the other layers' states. 

\subsection{Additional Output Layer Adaptation}\label{subsec:AdditionalUpdates}
Along with the threshold adaptation described in \cref{subsec:ClassificationSection}, additional weight update steps were investigated to speed up the convergence of learning. The first addition is the use of negative weight updates for misclassified spikes. If a neuron $n$ in the output layer belonging to a different class group than the label class group spikes for an input, the weights of the neuron are updated using a negative weight update according to \cref{eqn:negUpdate}.

\begin{align}
    W_{n} &= W_{n} - \eta * \Delta W_{n} \label{eqn:negUpdate}
\end{align}

The second additional weight update step investigated was rewarding the closest neuron in the label class group when there is no winner from this group. The closest neuron is the neuron with the highest dot product value ($v$ in \cref{eqn:membraneval}) among the neurons in the group corresponding to the label. The weights and threshold of the closest neuron $c$ with the highest dot product value among the label class group is updated according to \crefrange{eqn:closeUpdate}{eqn:CloseThreshUpdate}. This step is analogous to pulling the thresholds of the neurons in the label class group down until one of them spikes, and rewarding the first neuron that spikes.

\begin{align}
    W_{c} &= W_{c} + \eta * \Delta W_{c} \label{eqn:closeUpdate} \\
    \theta_{c} &= \theta_{c} + \eta_{thresh}*\Delta \theta_{c} \label{eqn:CloseThreshUpdate}
\end{align}
The threshold adaptation along with the two weight update steps together gave the best performance across all the tasks and the advantages in performance due to these weight update steps are discussed in \cref{subsec:weightUpdateComparison}

\section{Results}

We tested \acrshort{odesa} on different spatio-temporal transformation and classification tasks. The method was tested on a random pattern association task where random input spatio-temporal patterns were mapped to a target output spike stream. The network can simultaneously learn a hierarchical representation of the input patterns using an optimal number of neurons at each layer. \acrshort{odesa} was next tested on the IRIS dataset converted to precise temporal coded spike patterns using population coding proposed by Bohte et al. 2000 \cite{Bohte2000}. We also tested it on the Oxford spike pattern which was used to demonstrate the capabilities of SuperSpike \cite{zenke2018superspike}, as well as the latency-coded MNIST dataset. We then show the capabilities of the architecture by testing it on more complex problems like decoding Morse Code sequences and spoken digit classification using spikes from a Cochlea model. 
\begin{figure}[htbp]
\centering

  \includegraphics[width=0.7\linewidth]{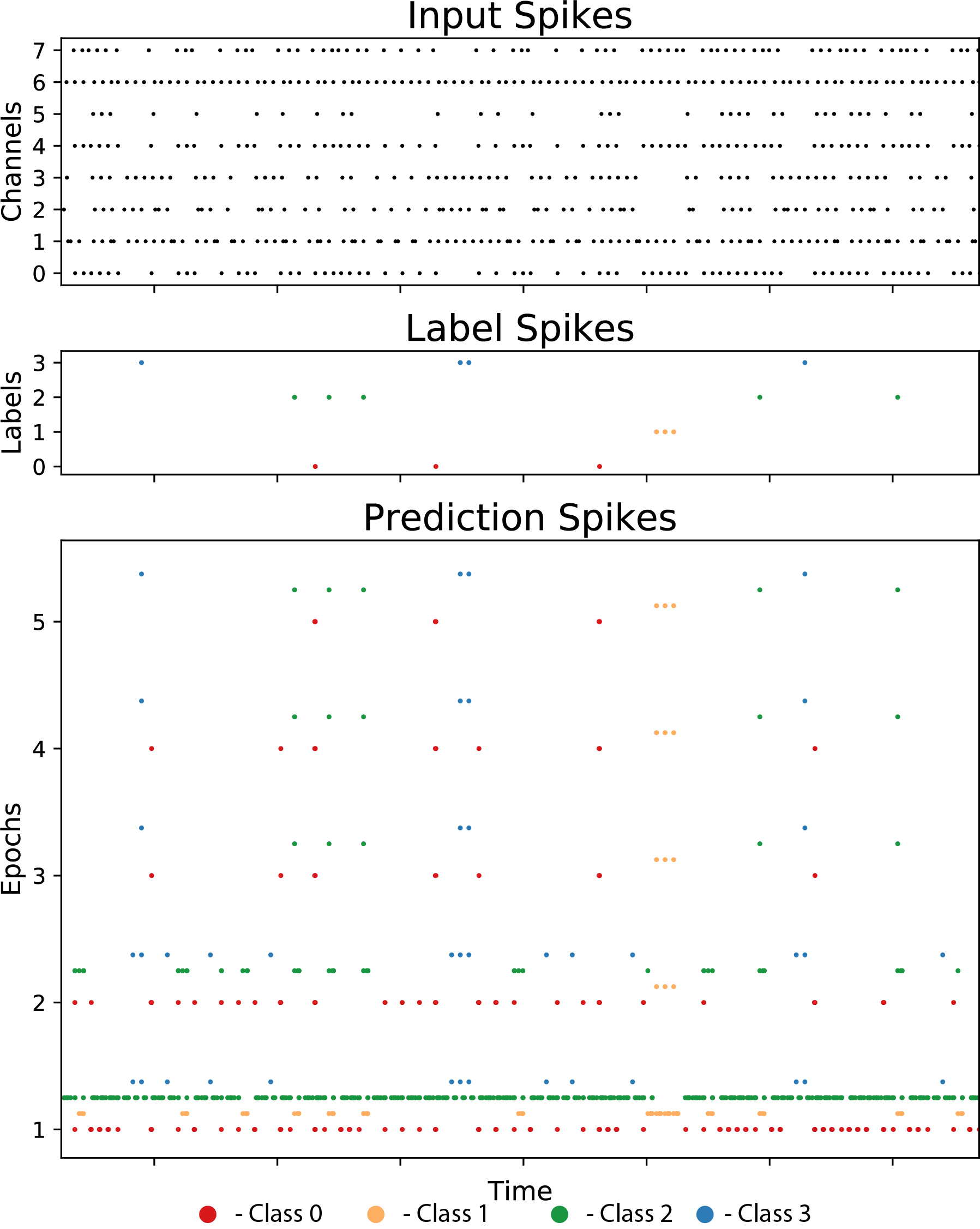}
\caption{Evolution of output layer activity for the random pattern association task. Embedded in the input spike patterns are unique sequences of symbols made of unique random patterns. The label spikes denote the desired output spike pattern for the given input spike pattern. The prediction spikes show the change in the activity of the 4 output neurons across 5 epochs.}
\label{fig:RandPat}

\end{figure}

\subsection{Random Pattern Association Task}
We first tested \acrshort{odesa} on a random pattern association task. Three random symbols (A, B, and C) were generated with 8 input channels that had a variable number of spikes per channel (6-8 spikes per symbol). We picked four target sequences randomly which are a combination of the constituent symbols (e.g., BBA, ACB, CAC, CCC). A stream of random combinations of the symbols ( "....B A A C B B C..." ) with no gap in between is presented to a two-layered \acrshort{odesa} network. A global attention signal is raised indicating the class at the last spike of each picked target sequence if and when they occur in the stream. The goal of the experiment is for the output layer of the \acrshort{odesa} network to generate spikes precisely at the time of the target label spikes. The network starts with spiking for every input event and gets rewarded every time it gets one of the label spikes right. The rewarded neurons get more specific in their spiking with time, and the neurons in the hidden layer get more specific in the features that occur just before the label spikes. \cref{fig:RandPat} shows the evolution of the output layer's activity across 1 to 5 epochs for a given random pattern association task. 

\subsection{Oxford Spike Pattern}
\begin{figure}[h]
\centering
\includegraphics[width=0.7\linewidth]{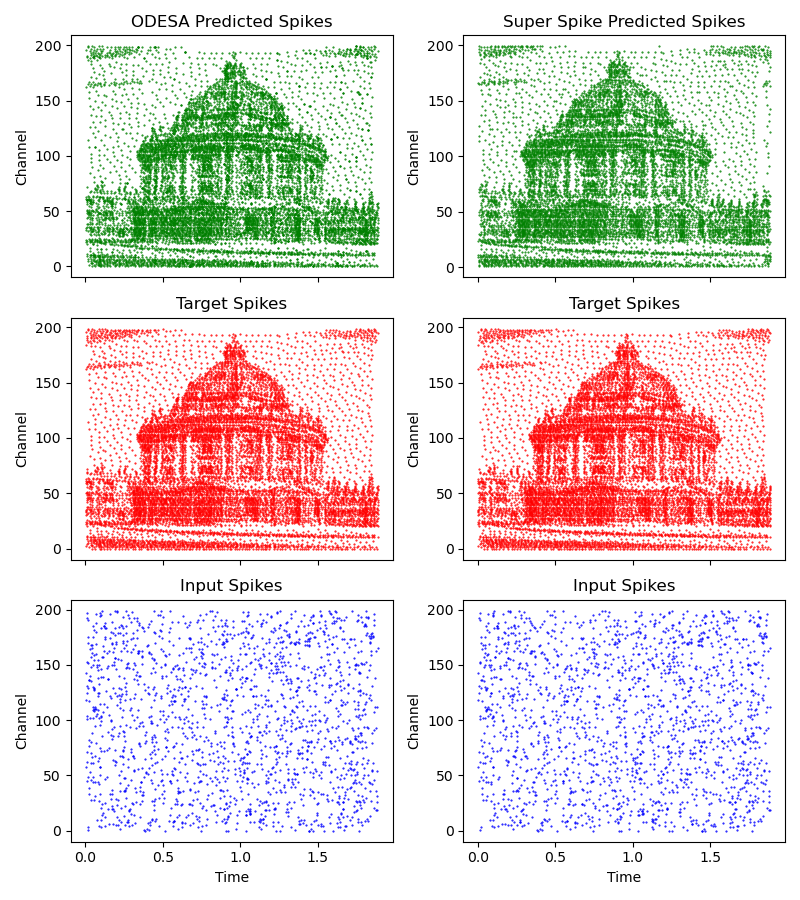}
\caption{Prediction of Oxford spike pattern using \acrshort{odesa} and the original SuperSpike algorithm using symmetric feedback.}
\label{fig:OxfordPrediction}

\end{figure}

 We also tested how the \acrshort{odesa} architecture can be adapted to other types of spike prediction problems such as the Oxford spike pattern, which was used to showcase the performance of the SuperSpike algorithm \cite{zenke2018superspike} in their supplementary code repository\footnote{https://github.com/fzenke/pub2018superspike}. The Oxford spike dataset consists of an input spike train, and a target spike train. The input spike train consists of random spikes generated in 200 channels over a period of 1.89 seconds. The target spike train is an image of a building that has been converted to a spatio-temporal spike train over 200 channels and 1.89 seconds. The task is to predict the precisely timed target spike train based on the random, but fixed, input spike train. This is very similar to the random pattern association task in the previous experiment, but with higher dimensions and without any inherent sub sequences in the input spike train. It is different from other datasets we tested \acrshort{odesa} in this work, due to the need to generate multiple output spikes per input spike. So the \acrshort{odesa} output layer was slightly modified to accommodate this requirement. We removed the hard \acrshort{wta} step in the output layer of the network and allowed multiple neurons to spike as long as their membrane values crossed their respective thresholds. The threshold adaptation of the output layer remained unchanged, and whenever a neuron group responsible for an output spike failed to spike, the thresholds were reduced. Neurons that correctly predicted the target spikes were rewarded. Also, as the \acrshort{odesa} architecture in its current form does not have delays in its synapses, at least one input spike should exist for every label spike. To facilitate this, we modified the target by mapping each spike in the output spike train to the nearest spike in the random input spike train. As the input spike train was fairly dense, the structure of the overall image did not change much from this process, as can be seen from the side by side comparison of target spikes in \cref{fig:OxfordPrediction}.
 
 \cref{fig:OxfordPrediction} shows the prediction from a single layer \acrshort{odesa} network and a two layer network trained with SuperSpike. The SuperSpike algorithm \cite{zenke2018superspike} uses an error measure that is a function of temporal difference between the predicted spike train and the target spike train. However, as \acrshort{odesa} makes predictions for each input spike, we had to create a different evaluation metric to monitor the training. For each input spike, we calculated the \acrfull{iou} over the sets of target spikes and predicted spikes. A mean \acrshort{iou} (between 0 and 1) per input spike was then calculated per epoch to evaluate the algorithm. With one output layer and no hidden layers, \acrshort{odesa} could achieve a mean \acrshort{iou} score of 0.80 for the Oxford spike pattern. The number of neurons per each correct class group ($k$) was proportional to the number of spikes in each target class. Though the receptive fields of neurons in \acrshort{odesa} can accommodate local temporal jitter, \acrshort{odesa} primarily learns patterns through clustering similar input patterns together. Therefore, it requires more than one neuron for each target class in the output layer to solve a problem like the Oxford spike pattern, as most of the features at each input spike are uncorrelated and unique. Two-thirds of the number of unique spikes in each target channel was used as the $k$ (number of neurons in each class group) value for that class group in the final network. That was the minimum number of neurons required in the output layer to predict most of the output spikes in our experiments. \cref{sec:deepOxfordSpike} discusses the performance of \acrshort{odesa} with different network depths on the Oxford spike pattern.

\subsection{IRIS Dataset}\label{subsec:iris}
To compare with other supervised learning methods on \acrshort{snn}s, we followed the same biologically plausible encoding scheme of traditional machine learning datasets as first proposed in Bohte et al. 2000 \cite{Bohte2000}.  The Fisher's IRIS dataset contains 3 classes with 50 samples each and is known for having linearly non-separable classes. Each sample has 4 input features, and $m = 5$ gaussian receptive fields per input feature were used to convert the features into $4 \times 5 = 20$ spiking channels. We used $\beta = 1.5$ just like the original work \cite{Bohte2000} where it was first used. We added no additional input channels to spike at intermediate intervals like used in \cite{taherkhani2018supervised}. Each input spiking channel emits a spike only once in the time between $t=0$ and $t=1$ secs per example. A simple 2-layer \acrshort{odesa} network was used with the hidden layer having time constant of $\tau_1 = 0.6 sec$ and an output layer with time constant of $\tau_2 = 0.9 sec$. The hidden layer had 10 neurons and the output layer had 1 neuron per class. We evaluated the network by performing 2-fold and 4-fold cross-validation. We compared the results with other supervised learning algorithms in \acrshort{snn}s and some non-spiking methods for reference in \cref{tab:iris}. We can see that \acrshort{odesa} can achieve comparable performance in the task with significantly fewer neurons and trainable parameters than other \acrshort{snn} training algorithms.
\begin{table}
\caption{Comparison of \acrshort{odesa} network on Fisher's IRIS Dataset with other methods}
    \centering
    \resizebox{\columnwidth}{!}{
\begin{tabular}{|c|c|c|c|c|c|c|}
    \hline
     Method & Inputs & Hidden & Outputs & Iterations & Train-Test Split & Testing Accuracy  \\
     \hline
     \multicolumn{7}{|c|}{\textit{Spiking Methods}} \\
     \hline
     RBF \cite{bohte2002unsupervised} & 32 & - & 3 & - & 50\%-50\% &92.6\%  \\
     SWAT \cite{wade2010swat} & 40 & 208 & 3 & 500 & 50\%-50\% &95.3\%  \\
     SpikeProp \cite{Bohte2000} & 50 & 10 & 1 & 1000 & 50\%-50\% &96.1\%  \\
     QuickProp \cite{ghosh2007improved} & 50 & 10 & 1 & 100-200 & 50\%-50\% &92.3\%  \\
     RProp \cite{ghosh2007improved} & 50 & 10 & 1 & Less than 100 & 50\%-50\% &93.2\%  \\
     RBF \cite{gueorguieva2006learning} & - & - & 3 & - & 50\%-50\% &89.0\%  \\
     Bako et al. \cite{sporea2013supervised} & 52 & - & 3 & 1000 & 50\%-50\% &83.4\%  \\
     TEM \cite{yu2014brain} & 48& - & 3 & Less than 100 & 50\%-50\% & 92.5\% \\
     Sporea et al. \cite{sporea2013supervised} & 48 & 200-300 & 3 & 100 & 75\%-25\% &94.0\%  \\
     Taherkhani et al. \cite{taherkhani2018supervised} & 169 & 360 & 3 & 100 & 75\%-25\% & 95.7\% \\
     \hline
     \hline
     Proposed Method & 20 & 10 & 3 & 200 & 75\%-25\% & 95.6\%\\
     Proposed Method & 20 & 10 & 3 & 200 & 50\%-50   \% & 95.1\%\\
      \hline
      \hline
     \multicolumn{7}{|c|}{\textit{Non-Spiking Methods}} \\
     \hline

     K-Means \cite{bohte2002unsupervised} & - & - & - & - & 50\%-50\% &88.6\%  \\
     SOM \cite{bohte2002unsupervised} & - & - & - & - & 50\%-50\% &85.3\%  \\
     MATLAB BP \cite{wade2010swat} & 50 & 10 & 3 & $2.6\times10^{6}$ & 50\%-50\% &95.7\%  \\
     \hline
\end{tabular}}

\label{tab:iris}
\end{table}

\subsection{Latency-Coded MNIST}

\begin{figure}
\centering

  \includegraphics[width=\linewidth]{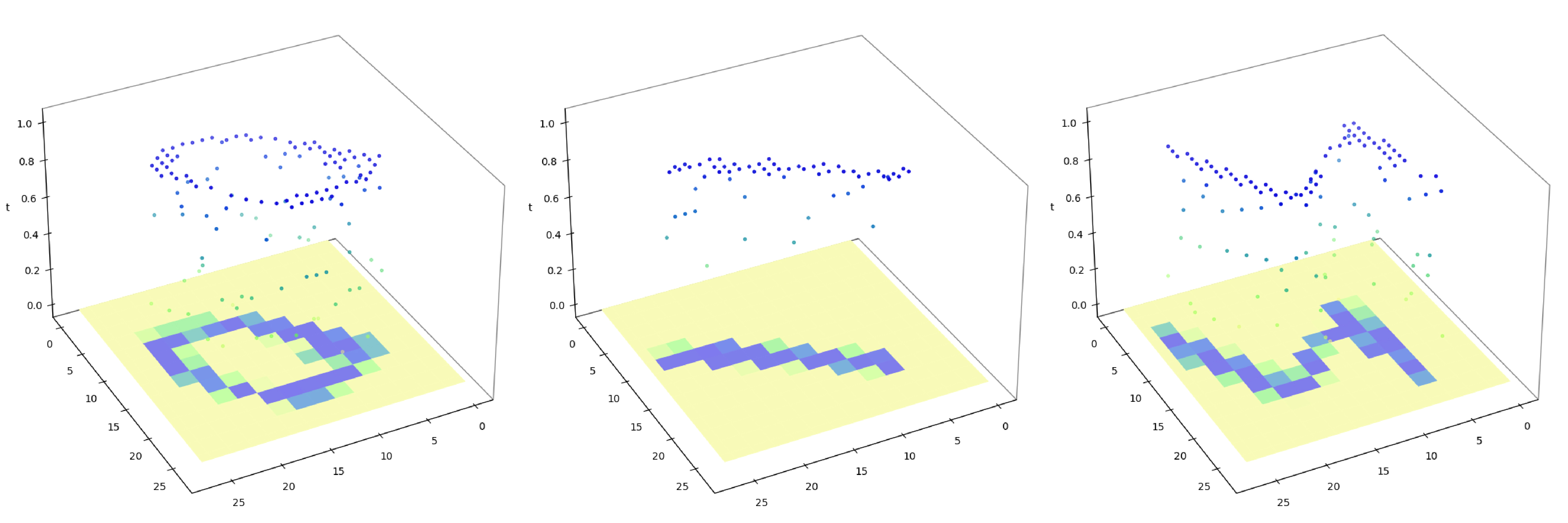}
\caption{Examples of classes '0','1',and '5' of latency coded spikes from MNIST}
\label{fig:mnistLatency}

\end{figure}
\begin{figure}
    \centering
    \begin{tabular}{cc}
      \includegraphics[width=0.5\linewidth]{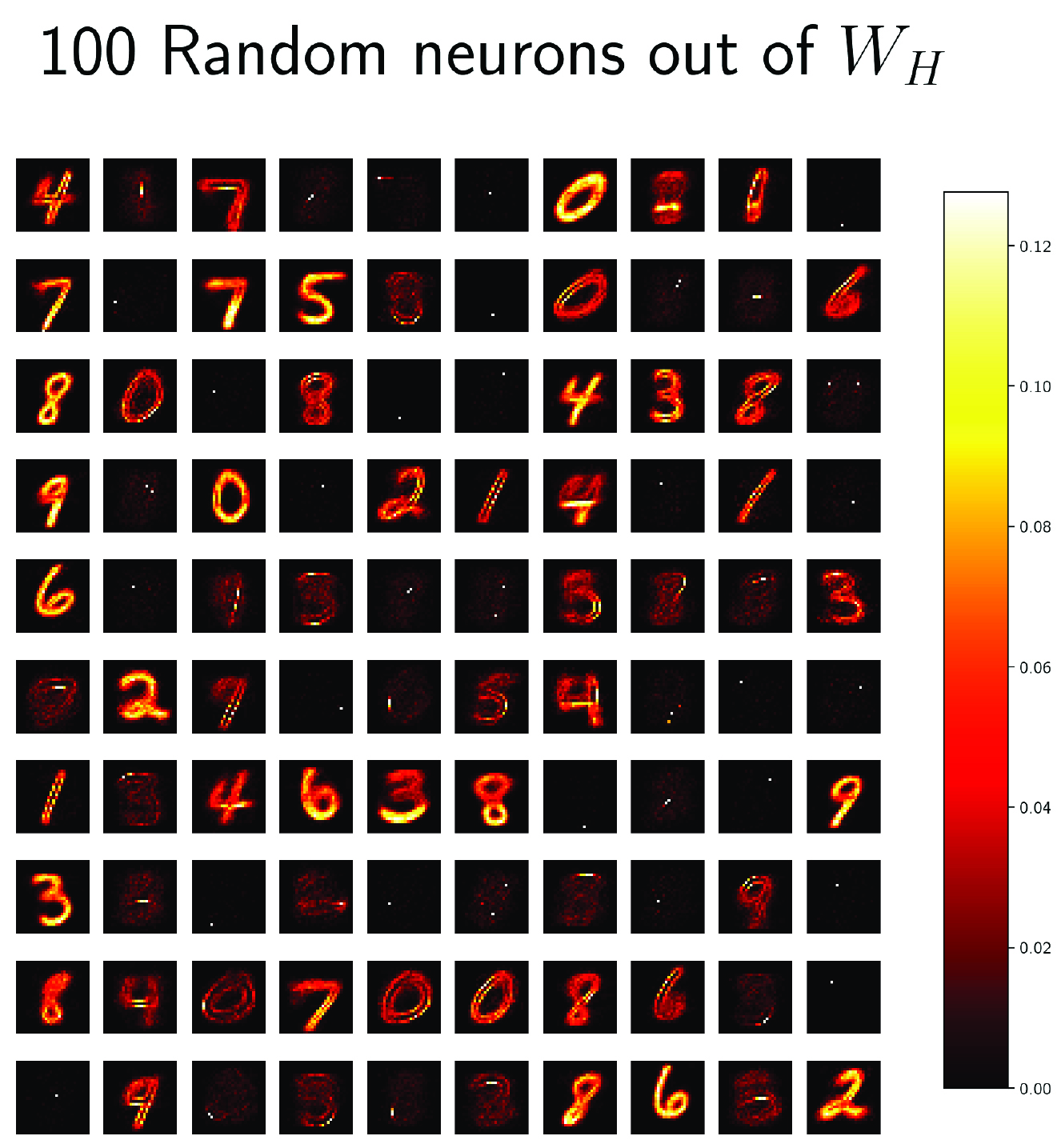}  &      \includegraphics[width=0.5\linewidth]{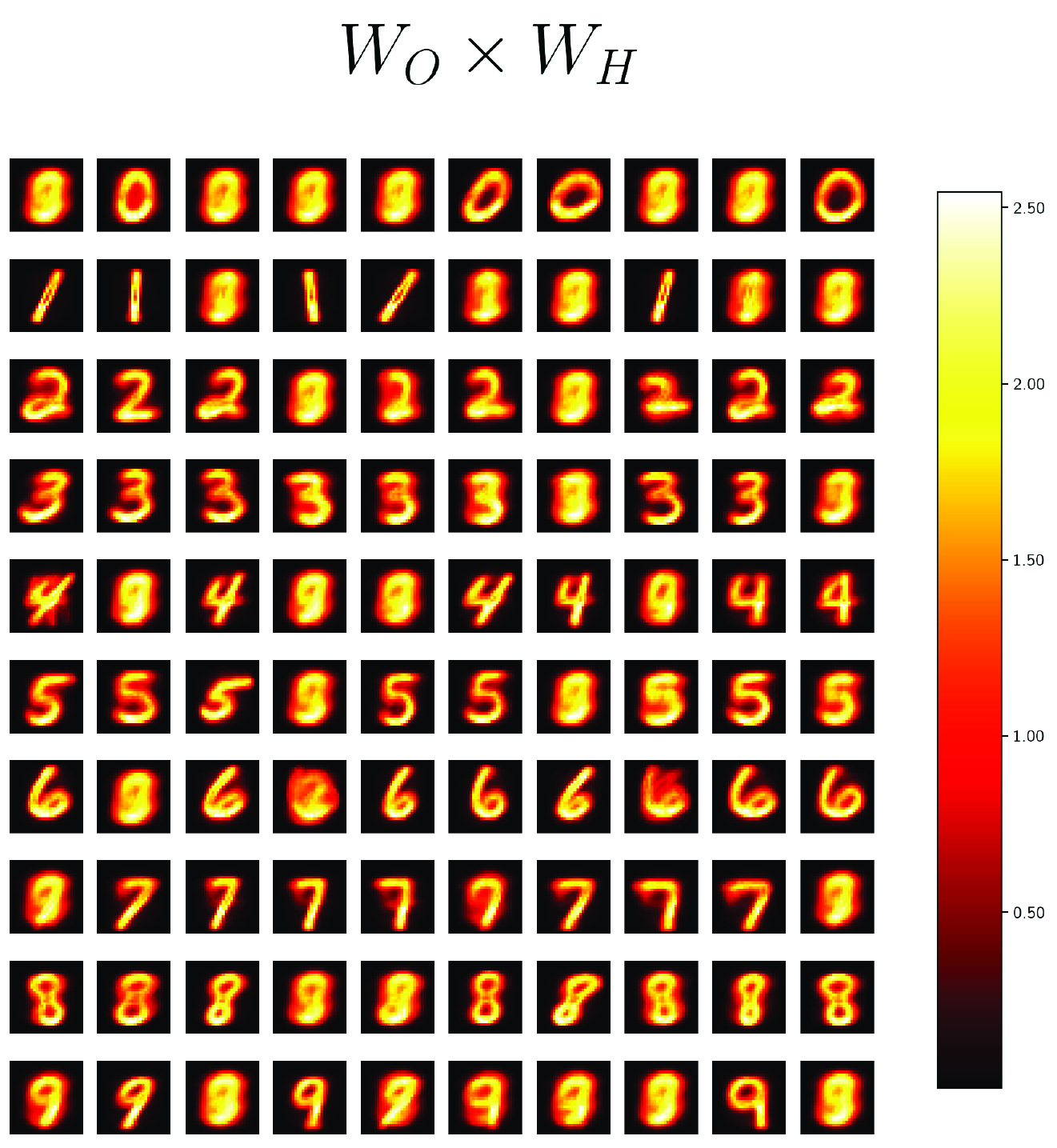}  \\
      (a) & (b) 
    \end{tabular}
    \caption{(a) Weights of 100 neurons randomly picked from hidden layer. (b) Composite weight visualisation of output neurons by multiplying the weight matrices of hidden layer and output layer. Each row represents the weights of neurons ($k = 10$) in each class group }
    \label{fig:mnistWeights}
\end{figure}

MNIST \cite{lecun1998gradient} consists of 60,000 training images belonging to 10 classes and 10,000 testing images. Each image is of size 28X28 which makes the input channels equal to 784. The MNIST images were preprocessed using latency coding to convert them into a spiking dataset. The brightness value of each pixel was linearly transformed from 0-255 to 0-1 seconds which was used as the timing of a spike from the corresponding input channel. All the spikes from pixels that had a timestamp of less than 0.3 seconds were eliminated to reduce the number of input spikes. As \acrshort{odesa} expects a precisely timed label spike, at the end of each example we generated a labelled spike that denotes the class of the example. The latency coding ensures that the input time surface context at the end of the example is similar to the original MNIST image. \cref{fig:mnistLatency} shows a few examples of input spikes generated this way. We have used an \acrshort{odesa} network with 1 hidden layer (6000 units) and 1 output layer ($k = 10$) to train on the latency-coded MNIST dataset to achieve a test accuracy of 93.23\%. The time constants for each layer $\tau_1$ and $\tau_2$ was set to $1.0sec$ each. We can easily visualise the features of time surface contexts learnt by the neurons in the hidden layer by simply plotting the heat map of the weights like shown in \cref{fig:mnistWeights} (a).

We can see that different neurons have learnt different intensity regions of the digits in MNIST, as the respective spikes fall close to each other temporally. We can also approximately estimate the patterns learnt by later layers by multiplying the weight matrices of a layer with its previous layer and so on. For example, in a two layer \acrshort{odesa} network, the composite weights of $2^{nd}$ layer can be estimated by a simple matrix multiplication (\cref{eq:compostiteWeights}). Where $W_2$ is the weight matrix of the $2^{nd}$ layer, and $W_1$ is the weight matrix of the $1^{st}$ layer. $n_i$, $n_1$, and $n_2$ represent the number of input channels, the number of neurons in $1^{st}$ layer, and the number of neurons in $2^{nd}$ layer respectively. It should be noted that this is only a limited 2D visualisation of the higher dimensional spatio-temporal feature learnt by the output neurons. Estimating the original time surface patterns which trigger the neurons is not easily tractable due to the weight normalisation and the non-linear thresholding operation.
\begin{equation}
    \underset{(n_2 \times n_i)}{W_{2}^{comp}} = \underset{(n_2 \times n_1)}{W_{2}} \cdot \underset{(n_1 \times n_i)}{W_{1}} \label{eq:compostiteWeights}
\end{equation}
\cref{fig:mnistWeights}(b) shows the composite weights of the output layer of \acrshort{odesa} network trained on MNIST. Each row in \cref{fig:mnistWeights}(b) represents the $k$ neurons in each class. The composite weights show that each neuron in an output class group is learning a different cluster of patterns for the class. 
\cref{tab:mnistComparison} shows the comparison of the test accuracy with various other \acrshort{snn} methods. \acrshort{odesa} performs on par with all the \acrshort{stdp}-based methods without requiring an additional classifier at the end.
 
\begin{table}
\caption{\label{tab:mnistComparison}Comparison of accuracy on the test set of MNIST with other \acrshort{snn} methods.}
\centering
\tiny
\resizebox{\columnwidth}{!}{
\begin{tabular}{|c|c|c|}

\hline
    Method  & Pre-Processing & Test Accuracy \\
    \hline
    \hline
    Two Layer with Exponential STDP +   & Rate Coding & 95.00\%\\
    \emph{a posteriori} neuron selection \cite{diehl2015unsupervised}& & \\
    \hline
    Spiking RBM +  & Thresholding +  & 89.00\% \\
    Contrastive Divergence, Linear Classifier \cite{merolla2011digital} & Rate coding & \\
    \hline
    Spiking RBM + & Thresholding + & 91.90\% \\
     Contrastive Divergence \cite{neftci2014event} &  Spike Coding  & \\
    \hline
    Dendritic neurons + & Rate Coding  & 90.30\% \\
    Morphology Learning \cite{hussain2014improved} & & \\
    \hline
    Multi-layer hierarchical network +  & Orientation Detection + & 91.60\% \\
    STDP with Calcium variable \cite{beyeler2013categorization} & Spike Coding & \\
    \hline
    Spiking CNN + & Scaling, Orientation, Thresholding + & 91.30\% \\
    Tempotron Rule \cite{zhao2014feedforward}& Spike Coding & \\
    \hline
    Three Layer SNN + & Rate Coding & 97.20\% \\
    STDP-Backpropagation \cite{tavanaei2019bp} & & \\
    \hline
    Deep SNN +  & Rate Coding & 98.60 \% \\
    Backpropagation \cite{lee2016training}& & \\
    \hline
    Two Layer SNN + & Latency based  & 96.92\% \\
    SGD \cite{mostafa2017supervised} & Spike Encoding & \\
    \hline
    ODESA (Our method) & Latency based   & 93.24\% \\
     & Spike Encoding& \\
     \hline
\end{tabular}}

\end{table}
\subsection{International Morse Code}

The problem with using the IRIS dataset and other machine learning datasets converted to spiking datasets using population encoding or latency coding is that each input channel can only spike once during the entire example. Furthermore, there is no hierarchy in the temporal features to be learnt. Hence, these datasets only test the spatial feature learning capabilities of an \acrshort{snn}.  Even popular neuromorphic vision datasets like N-MNIST do not have a hierarchy of temporal features, i.e., the time of the spike patterns is not key to the task, which is the classification of the digits such that the removal of all timing information (by simply binning all spikes at the input channel) does not result in any loss of information for the task. Thus, which spiking datasets test the true temporal learning capabilities of \acrshort{snn}s \cite{iyer2021neuromorphic} is still an open question. Models trained on the datasets like IRIS only learn one spike per channel, and don't have to learn a sequence of different states along the temporal dimension. Each example is a single spatio-temporal feature with no more than one spike per channel.  Hence, we made a custom task that can test and show the hierarchical learning capabilities of \acrshort{odesa} in the temporal dimension. We used the International Morse Code to encode different letters and numbers into spikes from two channels: "dash" and "dot". Each of these channels spikes multiple times for a given letter, and words would have multiple occurrences of the constituent letters. This forces the models trained on such a dataset to learn not only the spatio-temporal features across the channels at a given time but also learn the sequences of occurrences of such spatio-temporal features. For example, to differentiate two numerical sequences like "0,0,1,0,0" and "0,0,0,1,0", the model would have to learn the internal representation of the constituent numbers "0" and "1" which translate to "\textbf{- - - - -}", and "\textbf{. - - -}" in Morse code using dots (\textbf{.}) and dashes (\textbf{-}) as spikes. We generated multiple tasks which involve complex sequence learning to test the hierarchical spatio-temporal learning capabilities of \acrshort{odesa}.

The first task with Morse code was to classify four different names:  "ANDRE", "GREG", "SAEED", "YESH", and "YING" in Morse code. Each letter in each word was converted into a spike stream using the dots and dash encoding of Morse Code. For example, letter 'Y' is encoded as "\textbf{- . - -}" in Morse code and a spike is generated from the respective input channel for each dash '\textbf{-}' or dot '\textbf{.}'. A time gap was inserted between each letter to distinguish it from the next letter. At the last spike of the last letter of each word, a corresponding output class spike is generated in the output spike train. The task is to predict the output spike train for the sequence of input spikes that conveys the different names in the training set. We used a two-layered \acrshort{odesa} network, one layer to learn spatio-temporal features at the timescale of the letters, and the second layer to learn features at the timescale of words,  which is then fed into the output layer to learn the representations of the exact words in the dataset. 

\begin{figure}
    \centering
    \includegraphics[width=\linewidth]{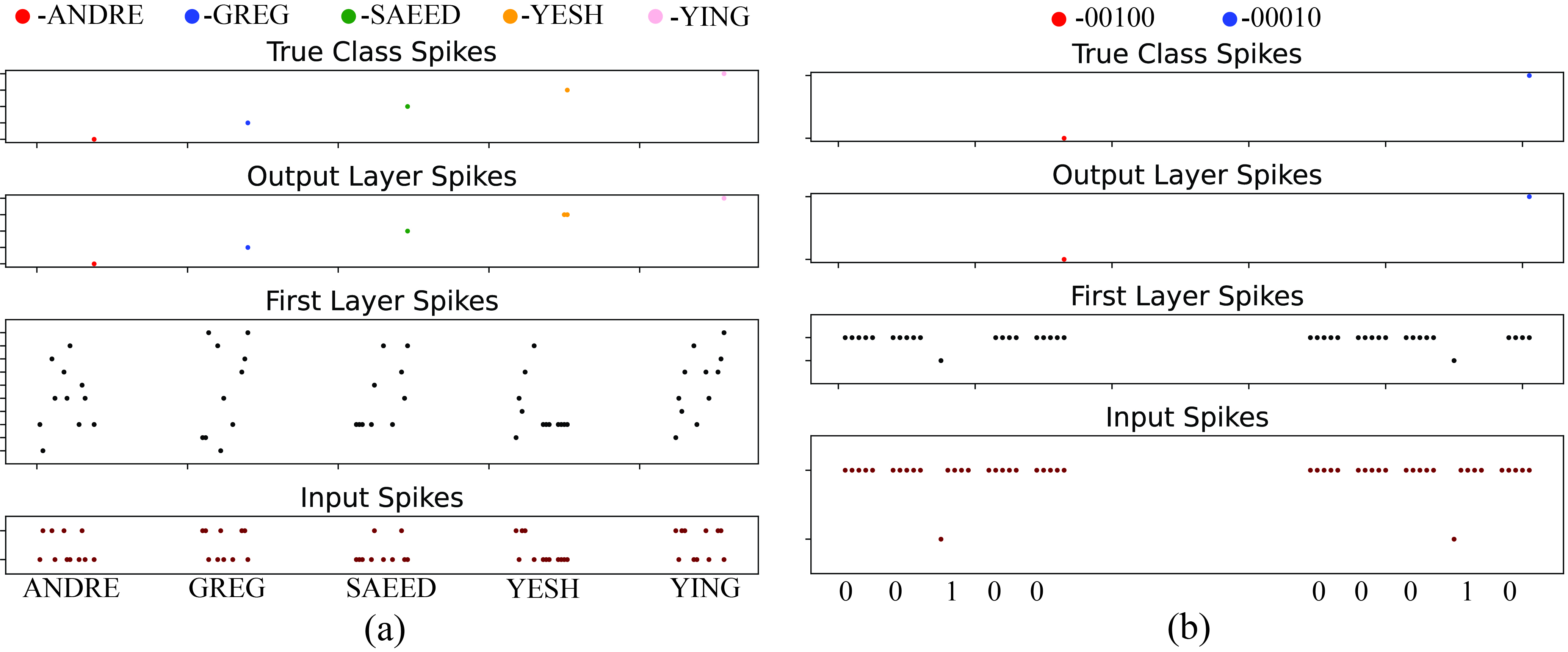}
    \caption{(a) Two-Layer \acrshort{odesa} activity after training to recognize the five author names "ANDRE", "GREG", "SAEED", "YESH", "YING" in Morse Code. (b)  Two-Layer \acrshort{odesa} activity after training to recognize "0,0,1,0,0" and "0,0,0,1,0" in Morse Code. }
    \label{fig:MorseDetectionTrained}
\end{figure}

The second task we tested \acrshort{odesa} on was the previously mentioned sequence of "0,0,1,0,0" and "0,0,0,1,0". Though the supervisory signal is only provided at the end of the last "0" in each sequence, the model would have to learn some intermediate representation of "1" to be able to solve the task. Along with learning the representation of the symbol "1", the network also has to learn the position of the same symbol in the sequence. We used a two-layered \acrshort{odesa} network, one layer to learn the symbols (0 and 1), and the second layer, which is the output layer, to learn the sequence of the symbols. The output layer has two neurons for the two classes. The \acrshort{odesa} network can learn an intermediate representation of the symbol "1" without relying on an explicit supervisory signal for symbol "1" due to the local supervision provided by the next layer. \cref{fig:MorseDetectionTrained} shows the network activity at each layer after training on both the digit sequence and name detection tasks.

\begin{figure}
\centering
  \includegraphics[width=\linewidth]{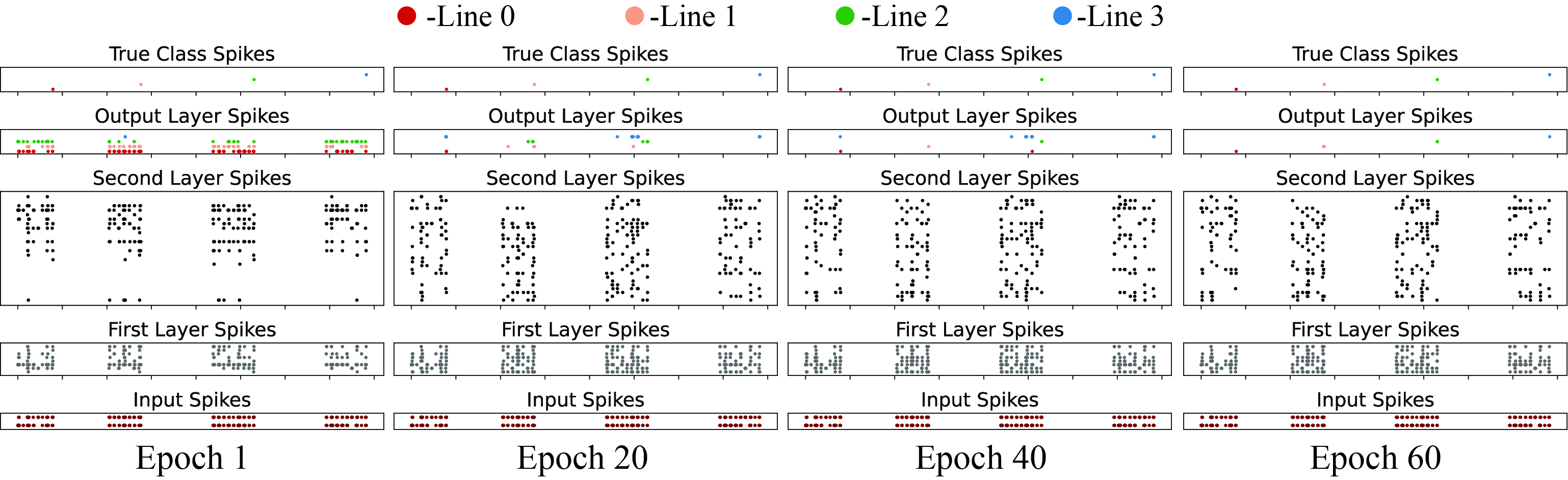}
\caption{Evolution of network activity over the epochs in training for the Sentence classification task}
\label{fig:activityevolution}
\end{figure}

\begin{figure}
\centering

  \includegraphics[width=0.7\linewidth]{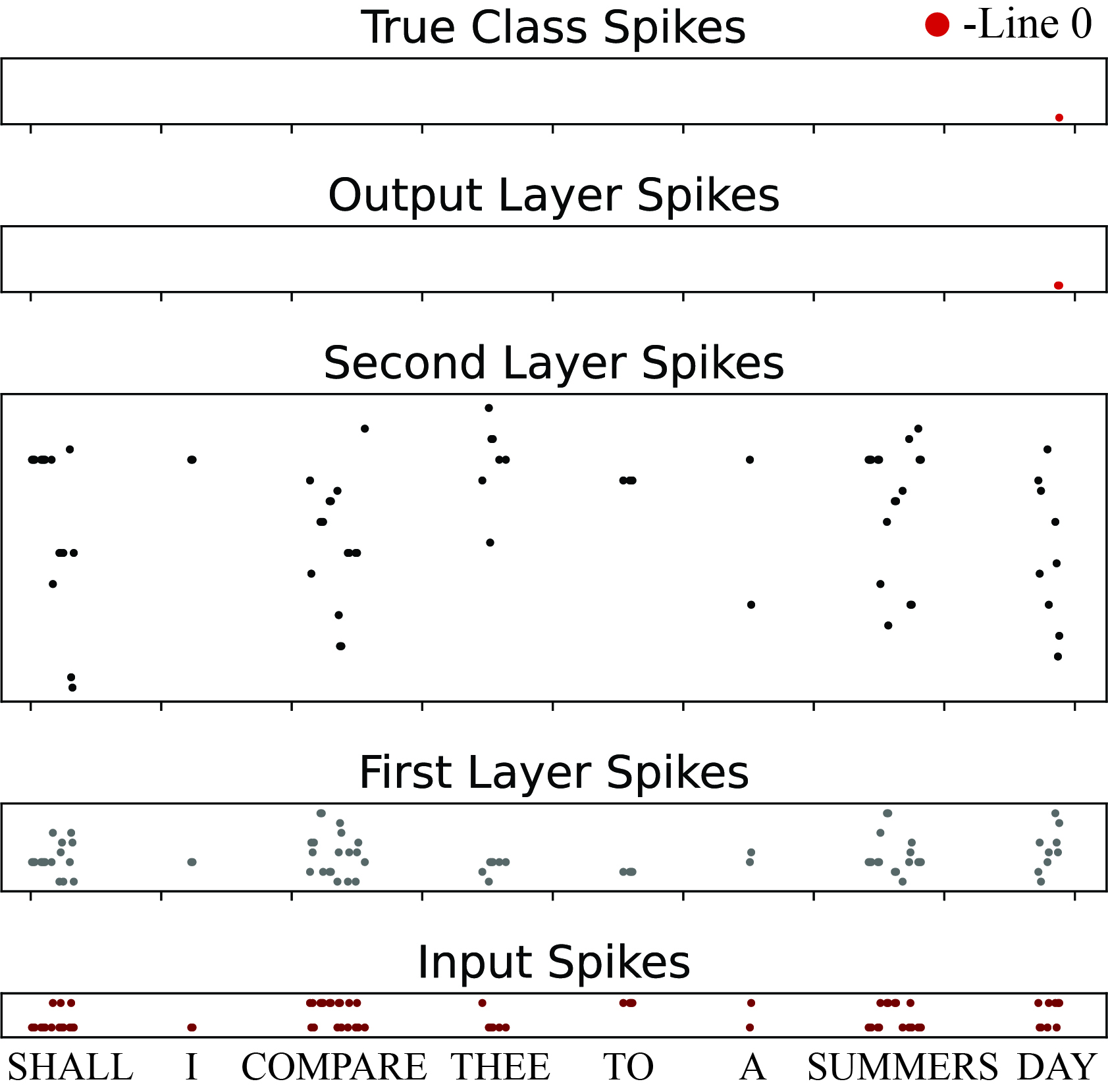}

\caption{Zoomed in network activity at each \acrshort{odesa} layer for the input line "SHALL I COMPARE THEE TO THE SUMMERS DAY" in Morse Code}
\label{fig:shakespearetrained}
\end{figure}

The third Morse Code task we tested \acrshort{odesa} on was a more complex sequence that involved learning sentences in Morse code. We took Shakespeare's  Sonnet 18 and used the first four lines as four different sequences to be learnt by the model. The four sentences were "\textbf{shall I compare thee to a summers day}", "\textbf{thou art more lovely and more temperate}", "\textbf{rough winds do shake the darling buds of may}", and "\textbf{and summer’s lease hath all too short a date}". The reason for selecting a poem for the sequence learning task was that poems have line endings that rhyme with each other. The last two letters of line 1 and line 3 are "ay", and the last three letters of line 2 and line 4 are "ate". This makes sure that the model cannot just learn the ending letters of each line to differentiate different lines. A three-layer \acrshort{odesa} network was trained for the task such that the first layer can learn the letters, the second layer can learn the words, and the third layer can learn the sentences. The time constant for the first layer was at the scale of the letters (5 timesteps), the second layer's time constant was at the scale of words (50 timesteps), and the output layer has a time constant appropriate for the scale of the sentences (200 timesteps). \cref{fig:activityevolution} shows the evolution of the activity at each layer at different epochs of the training. It shows how the specificity and the precise prediction of labels at the output layer with more training. \cref{fig:shakespearetrained} shows the final activity of the network for a single example after the training.

\subsection{TIDIGITS}

Next, we tested \acrshort{odesa} on the TIDIGITS corpus \cite{tidigits} for an isolated spoken digits recognition task. The TIDIGITS corpus includes isolated digits and digit sequences from both female and male speakers in different age groups. It thus provides sufficient speaker diversity. In \cite{Zhang_Wu_Chua_Luo_Pan_Liu_Li_2019}, the corpus was converted into a spike version using a threshold coding mechanism \cite{gutig2009time}. Each utterance from the corpus is firstly pre-processed by a 20-channel Constant-Q Transform (CQT) cochlear filter bank ranging from 200 Hz to 8 kHz. The generated spectrogram is then further encoded into spikes using the threshold coding.  For each cochlear output channel, 15 onset thresholds and 15 offset thresholds are set for the normalised amplitude. The upward and downward threshold crossing events of the channel represent an afferent to form a 30-afferent spike sequence. In this work, we use the generated spiking TIDIGITS dataset from \cite{Zhang_Wu_Chua_Luo_Pan_Liu_Li_2019} to test \acrshort{odesa}, and only isolated digits (11 classes) from adult female and male speakers are used, which includes 2464 digit utterances for training and 2486 for testing. MPD-AL \cite{zhang2019mpd} was used to classify the same spiking dataset using 10 neurons for each of the 11 classes in the dataset. MPD-AL is a version of aggregate-label learning which aims to achieve a desired spike count from a post-synaptic neuron based on the feedback signal. MPD-AL uses an iterative method to find the easiest modifiable time instant during the course of an input spike pattern based on the membrane potential traces of a neuron. The synaptic weights are then adjusted to add or remove post-synaptic spikes until the number of spikes matches the desired number of spikes. The 10 neurons assigned for each class are then trained to generate the desired number of spikes only for spike patterns that belong to their corresponding class and remain silent for other classes. Two different decoding schemes were used, based on the desired number of spikes to be generated, for the correct class neurons in \cite{zhang2019mpd}. The original work labelled the entire spike pattern as a single class and the number of spikes generated throughout an input spike pattern was used to determine the predicted class of the pattern. In our method, we labelled the last spike of each input spike pattern for a class with a label spike as \acrshort{odesa} requires a precisely timed label spike. We have used a three layer network with each layer learning spatio-temporal patterns at different timescales, i.e., 15~ms, 30~ms, and 35~ms respectively. The following table compares the results between the proposed method and MPD-AL. \cite{zhang2019mpd} on the same dataset. Unlike MPD-AL, ODESA networks do not access the entire history of the activity of a neuron during an input spike pattern. The supervisory label signal which indicates the class of an input spike pattern is only available to the neuron at the last spike in the spike pattern. It is possible to label every input spike with the label of the whole example, but that can often be misleading to the network, as two different digits can have similar sounds/phonemes in parts of the example. The framework of \acrshort{odesa} expects a correlation between the input pattern and the label spike. Hence we decided to only provide the labelled spike at the end of each example. In contrast, MPD-AL has access to the label of the input spike pattern at every instance of time. A three layer (2000-4000-11) \acrshort{odesa} network could achieve a training accuracy of 96.80\% and a  test accuracy of 91.4\% on the spiking TIDIGIT dataset. \cref{tab:tidigitComparison} shows the comparison of the performances with MPD-AL. 
\begin{table}
\caption{\label{tab:tidigitComparison}Test Accuracy results on TIDIGIT dataset.
}
\centering
\begin{tabular}{|c|c|}
\hline
    Method & Test Accuracy \\
    \hline
    MPD-AL with $N_{d} = 3$ \cite{zhang2019mpd}* & 95.3\%\\
    MPD-AL with Dynamic Decoding \cite{zhang2019mpd}* & 97.5\%\\
    \hline
    Proposed Method & 91.4\%\\
     \hline
\end{tabular}

\emph{(* The scores are reported as presented in \cite{zhang2019mpd}. However, they have not been verified as there was not enough information to reproduce and no code to do so was available.)}

\end{table}

\section{Discussion}
\subsection{Role of Weight Updates}\label{subsec:weightUpdateComparison}
As described in \cref{subsec:ClassificationSection}, threshold adaptation and weight update (\crefrange{eqn:OriginalWeightUpdate}{eqn:OriginalThreshUpdate}) in the output layer are the key steps that facilitate timely spiking of the neurons in the correct class group. The learning algorithm resulting from these updates in itself is capable of learning the features required to solve the benchmarks presented in this paper. However, the network failed to converge in some scenarios depending on the initial conditions. We investigated additional weight update steps for the output layer (\cref{subsec:AdditionalUpdates}) which can speed up the convergence of the network performance. One of the additional weight update steps was to use a negative weight update (\crefrange{eqn:closeUpdate}{eqn:CloseThreshUpdate}) which is similar to anti-\acrshort{stdp} to disincentivise neurons of wrong class groups from spiking. We observed that this step improved the final mean training accuracy on many datasets. The other weight update step investigated was rewarding the closest neuron in the correct class group in case none of the neurons in the correct class group spiked. This step improved the speed of convergence of the model more than the final accuracy. The closest winner update helps by rewarding the probable candidate output neuron in the correct class group immediately, without having to wait until the exact example reappear in the training data. 
\begin{figure}[h]
\centering
\includegraphics[width=\linewidth]{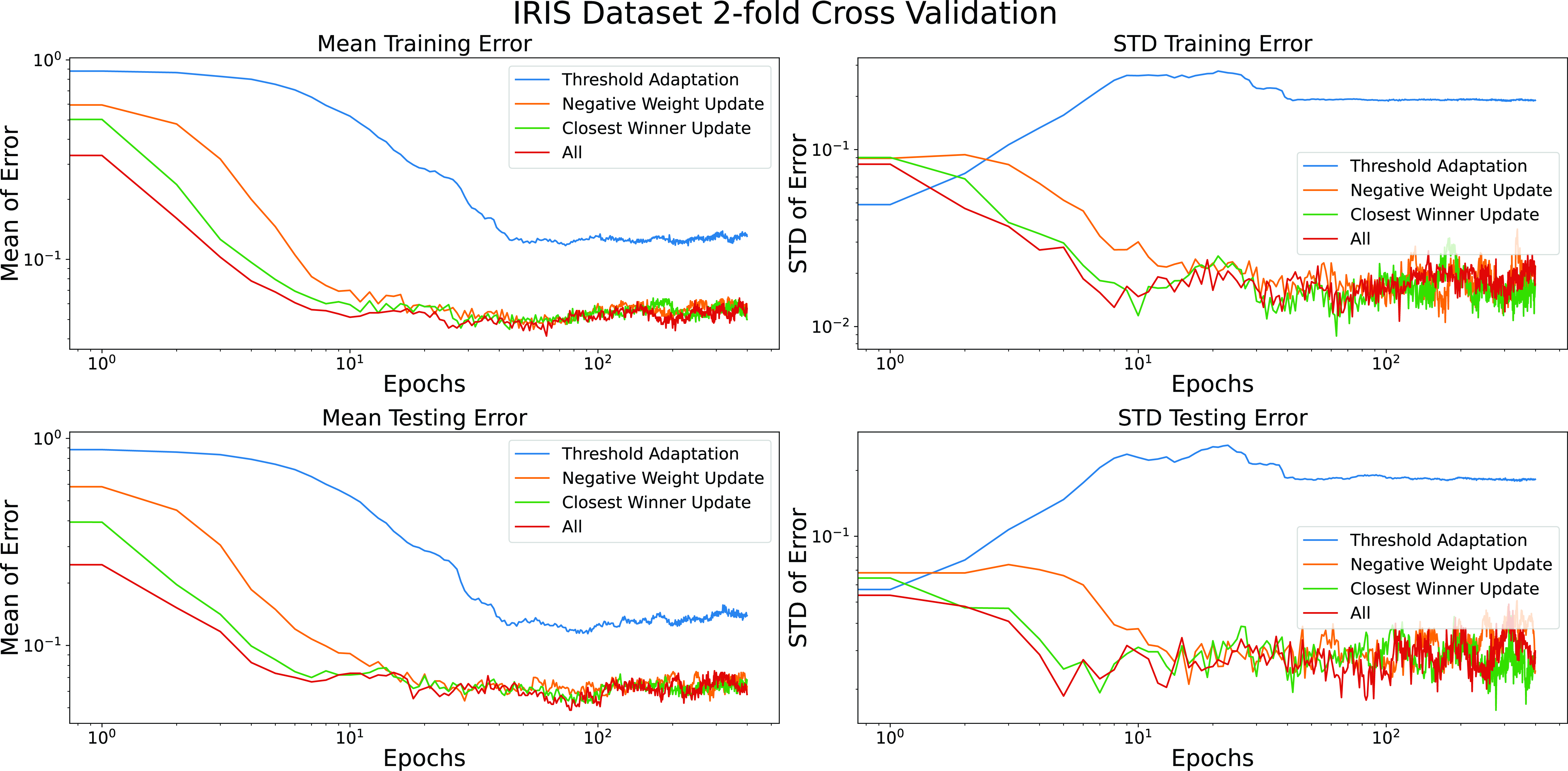}
\caption{Comparison of the model performance with different weight and threshold adaptation on the output layer}
\label{fig:variantComparison}
\end{figure}
Both the additional weight update steps individually improved the final accuracy of the models, and the combination of all the weight updates gave the best results in terms of the mean accuracy and its variance across multiple trials and initial conditions. We used the same IRIS dataset as in \cref{subsec:iris} to compare the effects of the additional weight update steps in improving the mean accuracy and the variance of the networks across 20 trials of 2-fold cross validation. \cref{fig:variantComparison} shows the mean of training and testing error across multiple trials at each epoch along with the standard deviation in the training and testing errors at each epoch. We don't yet fully understand why the models see a slight drop in the performance after reaching their peak and plan to investigate this in future work. We suspect some form of over-fitting as one of the reasons. 

\subsection{Deep learning in ODESA}\label{sec:deepOxfordSpike}
\begin{figure}[h]
\centering
\includegraphics[width=\linewidth]{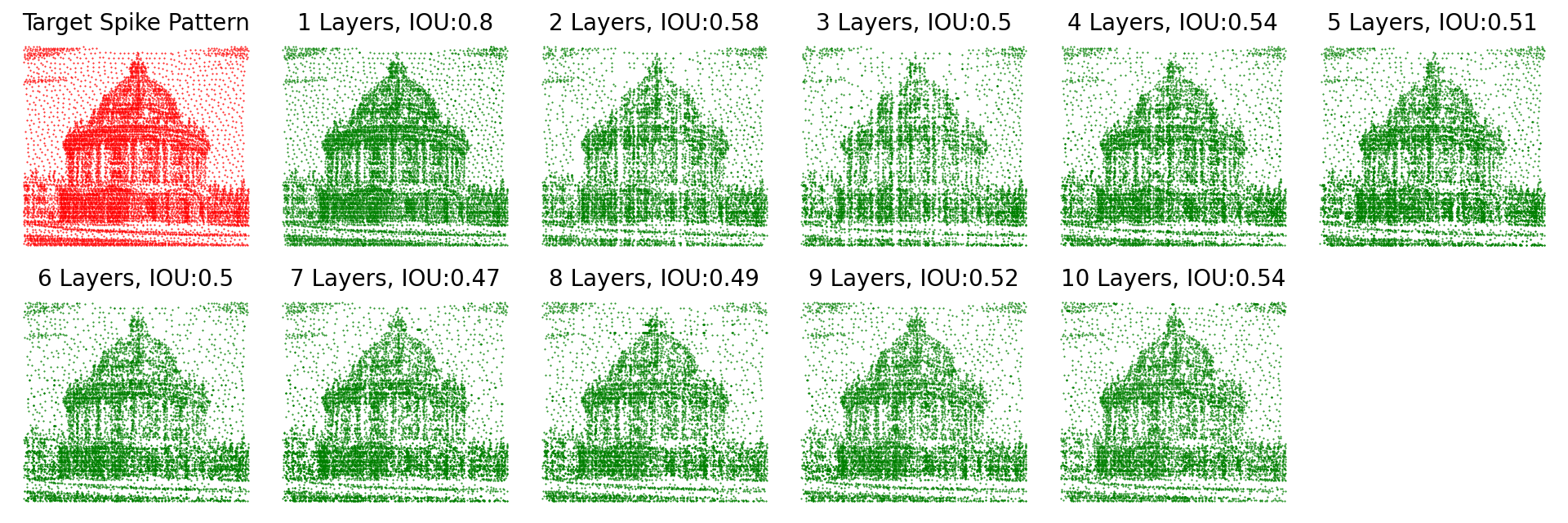}
\caption{Prediction of Oxford spike pattern from networks with different number of layers ranging from 1 to 10 layers including the output layer.}
\label{fig:DeepOxfordPerformance}
\end{figure}

\begin{figure}[h]
\centering
\includegraphics[width=0.8\linewidth]{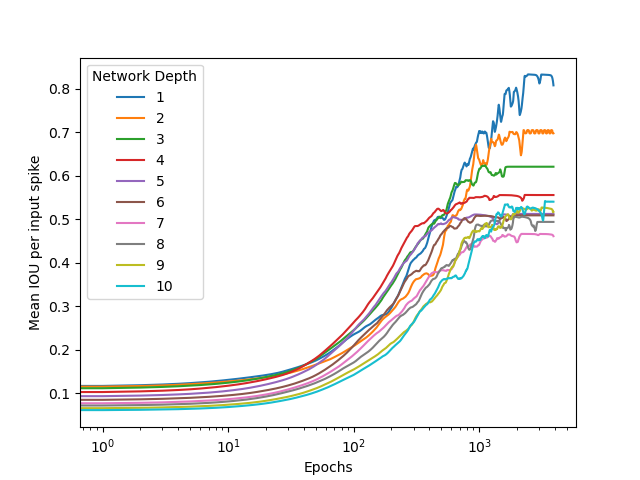}
\caption{Mean IOU per spike over Epochs for different depths of ODESA Networks on Oxford Spike Pattern }
\label{fig:DeepOxfordIOU}
\end{figure}

As the learning of \acrshort{odesa} inherently did not have any limitations on the depth of the models, we wanted to investigate the effect of deeper networks on the performance of the model. Unfortunately, none of the current spiking benchmarks available has enough temporal hierarchy to require much deeper networks that can utilise the hierarchy of temporal features. Nevertheless, we used the Oxford spike pattern to test any vanishing effects of the feedback attention signals used in \acrshort{odesa} with increasing depth of the models. We used networks of different depths from 1 layer to 10 layers (including the output layer) and all of them converged to a final accuracy after a sufficient number of training epochs. Different time constants ($\tau_{l}$) were used for each layer ranging from 0.001s to 0.0075s in the networks. The number of neurons also ranged from 300 to 1000 with the depth of the network. \cref{fig:DeepOxfordPerformance} shows the qualitative output of networks of different depths for the Oxford spike pattern. 
We have used the mean \acrshort{iou} per input spike as the measure of performance for the Oxford spike pattern. \cref{fig:DeepOxfordIOU} shows the trend of the mean \acrshort{iou} per input spike per epoch during the training of these networks. We observed that deeper networks required more epochs to converge to their final performance. This is expected because the latter layers of a network depend on the features of the earlier layers. We also noticed that there is a drop in performance when the networks get deeper, but there was no clear trend in this. Some deeper networks performed better than other shallower networks and vice versa. This can be due to multiple reasons including that each layer has to learn spatio-temporal features at different time scales depending on the time constant $\tau_l$ of the layers. Depending on the input spike data, the number of possible spatio-temporal features can be different at different time constants. These effects require a more detailed study to fully investigate. 

\subsection{Role of thresholds in ODESA}
There have been previous works that have used the neuronal spike thresholds in the learning process. Most commonly \acrshort{alif} neurons with dynamic thresholds have been shown to improve performance in recurrent \acrshort{snn}s \cite{bellec2018long} \cite{shaban2021adaptive} and promote homeostasis in the networks with \acrshort{stdp} \cite{diehl2015unsupervised}. Aggregate label learning methods like Multi-Spike Tempotron (MST)\cite{gutig2016spiking} and Threshold-Driven Plasticity (TDP)\cite{yu2018spike} used a Spike Threshold Surface to map the threshold of a neuron and the number of spikes it generates for a given spike train. Optimizing the weights using gradients with respect to the threshold enabled neurons to output the desired number of spikes per input pattern.  Alternatively, Membrane Potential Driven Aggregate Learning (MPD-AL) \cite{zhang2019mpd} used gradients with respect to the membrane potential, unlike MST and TDP. All the aggregate learning methods involve iterations to find the optimal threshold, or membrane voltage, to generate the desired number of spikes per neuron. But these learning algorithms cannot influence the precise timing of the spikes. Furthermore, the aggregate learning algorithms are single neuron algorithms and don't have solutions to learn hierarchical spatio-temporal features using multiple layers of neurons. Learning in \acrshort{odesa} doesn't involve finding the correct target threshold or membrane potential. Instead, \acrshort{odesa} finds the right weights and thresholds by online adaptation of the thresholds and facilitates neurons to spike precisely at the time of a global attention signal created by a supervisory signal. \acrshort{odesa} networks are capable of generating precisely timed spike patterns that can be learnt. In addition to that, \acrshort{odesa} can use spike-timing-dependent threshold adaptation to learn hierarchies of spatio-temporal features at multiple timescales by driving the adaptation of the previous layer's thresholds based on the current layer's activity. Evidence shows that rapid threshold adaptation in pre-synaptic neurons occurs with correlated spiking in the post-synaptic neuron as a form of spike-timing-dependent intrinsic plasticity in biology \cite{zhang2003other}.

\begin{figure}
\centering

  \includegraphics[width=0.7\linewidth]{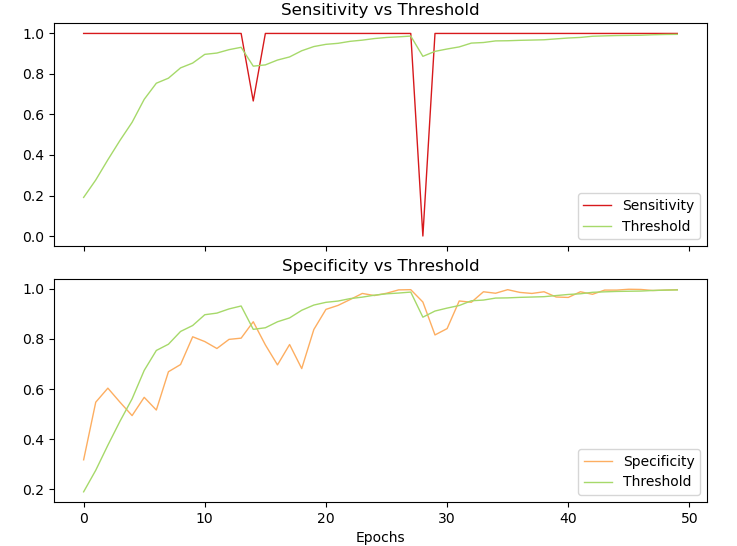}
\caption{The trend of Sensitivity and Specificity of a class in the Sentence classification task with respect to the threshold of the corresponding output neuron. Each epoch refers to 1200 time steps of the training simulation.}
\label{fig:sensvsspec}

\end{figure}

The thresholds in \acrshort{odesa} networks maintain neuronal homeostasis while also acting as gates to the adaptation of synaptic weights. When an output neuron does not spike for an important input spike, the threshold of the neuron is decreased until the neuron starts spiking. When any of those spikes match with the target spikes, the threshold of that neuron is increased while its weights are adjusted simultaneously. Higher thresholds make neurons more selective in their spiking. Every time the neuron weights are updated for an input time surface context, the likelihood of the neuron spiking again for the same input context increases, and the threshold increase makes it difficult for the neuron to spike for other contexts. If a neuron gets too specific, its thresholds are lowered again to let the neuron spike for a broader range of input spike patterns. This continues until the neuron reaches a balance between the desired activity and its activity. The desired activity for the output layer is the label output spike train.  Similarly, the goal of the hidden neurons is to support the spikes in the subsequent layer. Increasing the thresholds of the participant neurons and lowering the thresholds of inactive neurons for a spike in the next layer forces all the neurons in the layer to participate equally for every spike generated in the next layer. The combination of threshold adaptation and weight updates makes the neurons gradually become more specific in their spiking and match the spiking behaviour to the desired spike train. This can be seen in the evolution of layer activity throughout the training in \cref{fig:activityevolution}.

\cref{fig:sensvsspec} shows the trend of sensitivity and specificity of a class in the Sentence classification task. The drops in the threshold of an output neuron always occur with the drops in the sensitivity of a class. This is due to the punishment of the output neuron when it misses a ground truth label spike. An output neuron can miss the spikes when its threshold gets too high. On the other hand, the specificity of the class improves with a higher threshold of a neuron. When the threshold of a neuron is lowered, it also affects the specificity of the neuron as it spikes for other input spikes as well. This interplay between the rising and falling of the threshold helps the neuron to get more precise with its spiking and predict as many labelled spikes as possible. The threshold reaches the maximum value when no label spikes are missed and it precisely spikes only for the labelled input spikes. The thresholds can also be viewed as the confidence of a neuron for its corresponding class.

\subsection{Output Decoding in ODESA}
Different output decoding schemas are used in SNN models depending on output neurons using a Latency code or a Rate code. The time to first spike would require resetting to some default state from which the time of the spikes is supposed to be calculated. Rate coding on the other hand offers some error tolerance as a few missed spikes may not affect the firing rate of the output neurons. The output label spikes in ODESA are precisely timed with respect to the input spikes. The output spike from an ODESA model indicates the best prediction given all the evidence until that instant in time. ODESA performs the best when there exists a strong temporal correlation between the label spike and the labelled input spike. This sometimes can lead to an erroneous classification in some cases when the labelling is not precise. Tasks like Morse code detection, generally don't have this problem as they have a precise ending to a signal that can trigger the output classification. But there are very few openly available datasets that have temporally precise labelling. Spike-based TIDIGITS dataset examples usually have a few noisy input spikes at the end of each example which can make it challenging to learn the temporal correlation between the labelled input spike and the learnt label spike of the example.

\subsection{Advantages of ODESA}
Previous works have used multiple layers in \acrshort{snn}s to learn a spatial hierarchy of features at the same timescale. \acrshort{odesa} uses multi-layer networks to learn a spatio-temporal hierarchy of features. \acrshort{odesa} achieves sequence learning without having to rely on recurrent connections and only uses feed-forward connections in the network. This is achieved by calculating traces of neurons using longer time constants as we go deeper into the multi-layered network. Many of the previous event-driven feature extraction architectures \cite{sironi2018hats}\cite{lagorce2016hots}\cite{afshar2020event} also require a classifier at the the end of the \acrshort{snn} to achieve classification in \acrshort{snn}s. \acrshort{odesa} eliminates the requirement of such a separate classifier module and provides a solution using a spiking neural architecture from end to end. This architecture not only allows classification by the \acrshort{snn} but also allows them to learn arbitrary transformations between two spike streams. 

\acrshort{odesa} does not rely on surrogate gradients or real continuous-valued error feedback signals. The entirety of learning is achieved by only using binary attention signals, which are bio-plausible. This way, even when a deeper layer in the network goes silent, the earlier layers continue learning without having to wait for feedback signals from the silent layer. Another benefit of not using surrogate gradients and only using binary feedback signals is that \acrshort{odesa} doesn't have to rely on differentiable loss functions. In error back-propagation methods, the weights of a layer cannot be updated until an entire forward pass of prediction and backward pass of error calculation occurs. The learning of any hidden layer of \acrshort{odesa} only depends on the global attention signal from the labelled output spike train and the local attention signal from the layer above it. In essence, the learning of hidden layers in \acrshort{odesa} can continue even when a higher layer has gone completely silent. Furthermore, the computations in \acrshort{odesa} networks are all causal in the sense that no signals are propagated back in time, and no form of error \acrshort{bptt} occurs in the network. \acrshort{odesa} only uses computations that are event-driven and not clock-driven. This is valuable for energy-efficient neuromorphic hardware and can work well with neuromorphic sensors which may not generate data at all times. As the learning algorithm is online in nature, \acrshort{odesa} can also handle concept-drift in the data and keep updating its parameters in the light of any new changes in the input data distributions. But this can also lead to catastrophic forgetting in cases where the training data is not uniformly distributed.

The supervision in \acrshort{odesa} networks is not coupled with a fixed neuron model. The weight update steps in \cref{subsec:ClassificationSection} only perform the function of maximising the dot product of the winning input time surface context and the synaptic weights of the winner neuron. \acrshort{odesa} can be easily extended to other spiking neuronal models that allow threshold adaptation, with an equivalent synaptic weight update step that increases the likelihood of the neuron to spike again for the same input spike pattern. For neuron models with synaptic delays, the time constant of the trace of neurons has to be increased to take into account the delayed input to the next layer which may be crucial for future implementation in neuromorphic hardware. Similar modifications to the trace function (currently instantaneous exponential decay) would have to be made for higher-order dynamics of neuronal synaptic responses(e.g Alpha PSP). This opens door to more novel abstract spiking neuronal units that can be specialised to the task and input data (e.g., Neuromorphic vision data, Neuromorphic audio data, etc.). It also paves the way to use task-specific network architectures like convolutional layers and skip connections as used in ANNs which can induce task-specific priors to the features. In future work, the assumption of having an input spike for every ground truth spike can be relaxed by including delays in the neuron synaptic models.
\subsection{Limitations of ODESA}
The learning in each layer of \acrshort{odesa} is essentially a spike-based clustering algorithm similar to K-Means clustering. The neurons in \acrshort{odesa} learn features by learning the centres of the cluster of time surface contexts for which each neuron fires along with an acceptance boundary around this centre. Negative weights are not present in \acrshort{odesa} neurons due to the nature of the weight updates which are a form of the exponential moving average. Even the negative weight update step when added does not typically result in the generation of negative weights. The negative weight update step only helps in moving the weights away from the given time surface context in the high dimensional space. As can be seen in the \cref{fig:mnistWeights}, none of the weights is negative. This can potentially limit the performance compared to some other \acrshort{snn} methods which use neurons with negative weights. One possible solution is using an inhibitory neuron group that learns complimentary features using the same principles of \acrshort{odesa} in each layer. Alternate weight update paradigms which allow negative weights can also be one of the possibilities to explore in future work. Here we focused on extending the optimised \acrshort{snn} concept to multiple layers without resorting to error back-propagation which is not bio-plausible and challenging to implement in neuromorphic hardware.

\section{Conclusion}
In this work, we introduced a novel optimised abstraction to \acrshort{snn}s that can be used to solve practical machine learning problems in the spiking domain. We conducted a comprehensive evaluation of the architecture on various spiking datasets available. Unlike other methods that try to approximate error back-propagation to \acrshort{snn}s using techniques like surrogate gradients, we explore using a simple local spike-timing-dependent adaptation of thresholds and weights as the only way to train spiking neural architectures. The learning rule in \acrshort{odesa} networks is an event-driven supervised learning algorithm that enables learning transformations between arbitrary input and output spike trains under the single assumption that there exists a spike in the input train for every spike in the output spike train. We show that learning can be achieved in deep spiking neural architectures using binary attention signals between neuron groups and one global binary attention signal along with a simple three-factor adaptation of weights and thresholds. \acrshort{odesa} networks have sparse activity due to the hard \acrshort{wta} constraint on each layer, making it energy efficient to implement on neuromorphic hardware. This enables on-chip online learning capabilities for neuromorphic hardware. This work is an attempt at bridging the gap between machine learning and \acrshort{snn} models using a novel abstraction of biological spiking networks without relying on back-propagation of gradients.

\section{Author Contributions}
Yeshwanth Bethi, Saeed Afshar, and Andr\'e van Schaik conceived the overall theory and design of the algorithm. Yeshwanth Bethi, Saeed Afshar, and Ying Xu performed the experiments. Yeshwanth Bethi wrote the initial draft of the manuscript. Saeed Afshar, Andr\'e van Schaik, Ying Xu, and Greg Cohen edited and provided feedback on the manuscript.

\section{Acknowledgement}
This research was supported by the Commonwealth of Australia through the NGTF Cyber Call 2019 and in collaboration with the Defence Science and Technology Group of the Department of Defence.

\bibliographystyle{unsrt}

\begin{thebibliography}{10}

    \bibitem{rumelhart1986learning}
    David~E Rumelhart, Geoffrey~E Hinton, and Ronald~J Williams.
    \newblock {Learning representations by back-propagating errors}.
    \newblock {\em Nature}, 323(6088):533--536, 1986.
    
    \bibitem{Stork1989}
    David~G. Stork.
    \newblock {Is backpropagation biologically plausible?}
    \newblock In {\em International Joint Conference on Neural Networks}, pages
      241--246, 1989.
    
    \bibitem{crick1989recent}
    Francis Crick.
    \newblock {The recent excitement about neural networks.}
    \newblock {\em Nature}, 337(6203):129--132, 1989.
    
    \bibitem{grossberg1987competitive}
    Stephen Grossberg.
    \newblock {Competitive learning: From interactive activation to adaptive
      resonance}.
    \newblock {\em Cognitive Science}, 11(1):49--50, 1987.
    
    \bibitem{Bohte2000}
    S~M Bohte, H~{La Poutr{\'{e}}}, and J~N Kok.
    \newblock {Error-Backpropagation in Temporally Encoded Networks of Spiking
      Neurons}.
    \newblock {\em Neurocomputing}, 48:17--37, 2000.
    
    \bibitem{gutig2006tempotron}
    Robert G{\"{u}}tig and Haim Sompolinsky.
    \newblock {The tempotron: a neuron that learns spike timing--based decisions}.
    \newblock {\em Nature Neuroscience}, 9(3):420--428, 2006.
    
    \bibitem{florian2012chronotron}
    R{\u{a}}zvan~V Florian.
    \newblock {The chronotron: A neuron that learns to fire temporally precise
      spike patterns}.
    \newblock {\em PLoS ONE}, 2012.
    
    \bibitem{Neftci2019}
    Emre~O. Neftci, Hesham Mostafa, and Friedemann Zenke.
    \newblock {Surrogate Gradient Learning in Spiking Neural Networks: Bringing the
      Power of Gradient-based optimization to spiking neural networks}.
    \newblock {\em IEEE Signal Processing Magazine}, 36(6):51--63, nov 2019.
    
    \bibitem{bellec2018long}
    Guillaume Bellec, Darjan Salaj, Anand Subramoney, Robert Legenstein, and
      Wolfgang Maass.
    \newblock {Long short-term memory and learning-to-learn in networks of spiking
      neurons}.
    \newblock {\em Advances in Neural Information Processing Systems}, 31:795--805,
      2018.
    
    \bibitem{bellec2020solution}
    Guillaume Bellec, Franz Scherr, Anand Subramoney, Elias Hajek, Darjan Salaj,
      Robert Legenstein, and Wolfgang Maass.
    \newblock {A solution to the learning dilemma for recurrent networks of spiking
      neurons}.
    \newblock {\em Nature Communications}, 11(1):1--15, 2020.
    
    \bibitem{zenke2018superspike}
    Friedemann Zenke and Surya Ganguli.
    \newblock {Superspike: Supervised learning in multilayer spiking neural
      networks}.
    \newblock {\em Neural Computation}, 30(6):1514--1541, 2018.
    
    \bibitem{zipser1988back}
    David Zipser and Richard~A Andersen.
    \newblock {A back-propagation programmed network that simulates response
      properties of a subset of posterior parietal neurons}.
    \newblock {\em Nature}, 331(6158):684, 1988.
    
    \bibitem{mcclelland1995there}
    James~L McClelland, Bruce~L McNaughton, and Randall~C O'Reilly.
    \newblock {Why there are complementary learning systems in the hippocampus and
      neocortex: insights from the successes and failures of connectionist models
      of learning and memory.}
    \newblock {\em Psychological Review}, 102(3):439, 1995.
    
    \bibitem{zenke2021remarkable}
    Friedemann Zenke and Tim~P Vogels.
    \newblock {The remarkable robustness of surrogate gradient learning for
      instilling complex function in spiking neural networks}.
    \newblock {\em Neural Computation}, 33(4):899--925, 2021.
    
    \bibitem{lillicrap2016random}
    Timothy~P Lillicrap, Daniel Cownden, Douglas~B Tweed, and Colin~J Akerman.
    \newblock {Random synaptic feedback weights support error backpropagation for
      deep learning}.
    \newblock {\em Nature Communications}, 7(1):1--10, 2016.
    
    \bibitem{tavanaei2016acquisition}
    Amirhossein Tavanaei, Timoth{\'{e}}e Masquelier, and Anthony~S Maida.
    \newblock {Acquisition of visual features through probabilistic
      spike-timing-dependent plasticity}.
    \newblock In {\em 2016 International Joint Conference on Neural Networks
      (IJCNN)}, pages 307--314. IEEE, 2016.
    
    \bibitem{Mozafari2018}
    Milad Mozafari, Saeed~Reza Kheradpisheh, Timothee Masquelier, Abbas
      Nowzari-Dalini, and Mohammad Ganjtabesh.
    \newblock {First-spike-based visual categorization using reward-modulated
      STDP}.
    \newblock {\em IEEE Transactions on Neural Networks and Learning Systems},
      29(12):6178--6190, 2018.
    
    \bibitem{vigneron2020critical}
    Alex Vigneron and Jean Martinet.
    \newblock {A critical survey of STDP in Spiking Neural Networks for Pattern
      Recognition}.
    \newblock In {\em International Joint Conference on Neural Networks (IJCNN)},
      pages 1--9. IEEE, 2020.
    
    \bibitem{paredes2019unsupervised}
    Federico Paredes-Vall{\'{e}}s, Kirk Y~W Scheper, and Guido C H~E de~Croon.
    \newblock {Unsupervised learning of a hierarchical spiking neural network for
      optical flow estimation: From events to global motion perception}.
    \newblock {\em IEEE Transactions on Pattern Analysis and Machine Intelligence},
      42(8):2051--2064, 2019.
    
    \bibitem{legenstein2005can}
    Robert Legenstein, Christian Naeger, and Wolfgang Maass.
    \newblock {What can a neuron learn with spike-timing-dependent plasticity?}
    \newblock {\em Neural Computation}, 17(11):2337--2382, 2005.
    
    \bibitem{ponulak2010supervised}
    Filip Ponulak and Andrzej Kasi{\'{n}}ski.
    \newblock {Supervised learning in spiking neural networks with ReSuMe: sequence
      learning, classification, and spike shifting}.
    \newblock {\em Neural Computation}, 22(2):467--510, 2010.
    
    \bibitem{widrow1960adaptive}
    Bernard Widrow and Marcian~E Hoff.
    \newblock {Adaptive switching circuits}.
    \newblock Technical report, Stanford Univ Ca Stanford Electronics Labs, 1960.
    
    \bibitem{mohemmed2012span}
    Ammar Mohemmed, Stefan Schliebs, Satoshi Matsuda, and Nikola Kasabov.
    \newblock {Span: Spike pattern association neuron for learning spatio-temporal
      spike patterns}.
    \newblock {\em International Journal of Neural Systems}, 22(04):1250012, 2012.
    
    \bibitem{sporea2013supervised}
    Ioana Sporea and Andr{\'{e}} Gr{\"{u}}ning.
    \newblock {Supervised learning in multilayer spiking neural networks}.
    \newblock {\em Neural Computation}, 25(2):473--509, 2013.
    
    \bibitem{taherkhani2018supervised}
    Aboozar Taherkhani, Ammar Belatreche, Yuhua Li, and Liam~P Maguire.
    \newblock {A supervised learning algorithm for learning precise timing of
      multiple spikes in multilayer spiking neural networks}.
    \newblock {\em IEEE Transactions on Neural Networks and Learning Systems},
      29(11):5394--5407, 2018.
    
    \bibitem{schliebs2013evolving}
    Stefan Schliebs and Nikola Kasabov.
    \newblock {Evolving spiking neural network—a survey}.
    \newblock {\em Evolving Systems}, 4(2):87--98, 2013.
    
    \bibitem{wysoski2006adaptive}
    Simei~Gomes Wysoski, Lubica Benuskova, and Nikola Kasabov.
    \newblock {Adaptive learning procedure for a network of spiking neurons and
      visual pattern recognition}.
    \newblock In {\em International Conference on Advanced Concepts for Intelligent
      Vision Systems}, pages 1133--1142. Springer, 2006.
    
    \bibitem{Lobo2018}
    Jesus~L. Lobo, Ibai La{\~{n}}a, Javier {Del Ser}, Miren~Nekane Bilbao, and
      Nikola Kasabov.
    \newblock {Evolving Spiking Neural Networks for online learning over drifting
      data streams}, 2018.
    
    \bibitem{Kasabov2019}
    Nikola~K Kasabov.
    \newblock Evolving spiking neural networks.
    \newblock In {\em Time-space, spiking neural networks and brain-inspired
      artificial intelligence}, pages 169--199. Springer, 2019.
    
    \bibitem{belatreche2006evolutionary}
    Ammar Belatreche, Liam~P Maguire, Martin McGinnity, and Qing~Xiang Wu.
    \newblock {Evolutionary design of spiking neural networks}.
    \newblock {\em New Mathematics and Natural Computation}, 2(03):237--253, 2006.
    
    \bibitem{Pavlidis2005}
    N.~G. Pavlidis, D.~K. Tasoulis, V.~P. Plagianakos, G.~Nikiforidis, and M.~N.
      Vrahatis.
    \newblock {Spiking neural network training using evolutionary algorithms}.
    \newblock In {\em Proceedings of the International Joint Conference on Neural
      Networks}, volume~4, pages 2190--2194, 2005.
    
    \bibitem{Vazquez2011}
    Roberto~A. Vazquez.
    \newblock {Training spiking neural models using cuckoo search algorithm}.
    \newblock In {\em IEEE Congress of Evolutionary Computation}, pages 679--686,
      2011.
    
    \bibitem{wang2015spiketemp}
    Jinling Wang, Ammar Belatreche, Liam~P Maguire, and Thomas~Martin McGinnity.
    \newblock {Spiketemp: An enhanced rank-order-based learning approach for
      spiking neural networks with adaptive structure}.
    \newblock {\em IEEE Transactions on Neural Networks and Learning Systems},
      28(1):30--43, 2015.
    
    \bibitem{gutig2016spiking}
    Robert G{\"{u}}tig.
    \newblock {Spiking neurons can discover predictive features by aggregate-label
      learning}.
    \newblock {\em Science}, 351(6277), 2016.
    
    \bibitem{yu2018spike}
    Qiang Yu, Haizhou Li, and Kay~Chen Tan.
    \newblock {Spike timing or rate? neurons learn to make decisions for both
      through threshold-driven plasticity}.
    \newblock {\em IEEE Transactions on Cybernetics}, 49(6):2178--2189, 2018.
    
    \bibitem{zhang2019mpd}
    Malu Zhang, Jibin Wu, Yansong Chua, Xiaoling Luo, Zihan Pan, Dan Liu, and
      Haizhou Li.
    \newblock {MPD-AL: an efficient membrane potential driven aggregate-label
      learning algorithm for spiking neurons}.
    \newblock In {\em Proceedings of the AAAI Conference on Artificial
      Intelligence}, volume~33, pages 1327--1334, 2019.
    
    \bibitem{farries2007reinforcement}
    Michael~A Farries and Adrienne~L Fairhall.
    \newblock {Reinforcement learning with modulated spike timing--dependent
      synaptic plasticity}.
    \newblock {\em Journal of Neurophysiology}, 98(6):3648--3665, 2007.
    
    \bibitem{brea2013matching}
    Johanni Brea, Walter Senn, and Jean-Pascal Pfister.
    \newblock {Matching recall and storage in sequence learning with spiking neural
      networks}.
    \newblock {\em Journal of Neuroscience}, 33(23):9565--9575, 2013.
    
    \bibitem{friedrich2011spatio}
    Johannes Friedrich, Robert Urbanczik, and Walter Senn.
    \newblock {Spatio-temporal credit assignment in neuronal population learning}.
    \newblock {\em PLoS Computational Biology}, 7(6):e1002092, 2011.
    
    \bibitem{florian2007reinforcement}
    R{\u{a}}zvan~V Florian.
    \newblock {Reinforcement learning through modulation of spike-timing-dependent
      synaptic plasticity}.
    \newblock {\em Neural Computation}, 19(6):1468--1502, 2007.
    
    \bibitem{jimenez2014stochastic}
    Danilo {Jimenez Rezende} and Wulfram Gerstner.
    \newblock {Stochastic variational learning in recurrent spiking networks}.
    \newblock {\em Frontiers in Computational Neuroscience}, 8:38, 2014.
    
    \bibitem{fremaux2016neuromodulated}
    Nicolas Fr{\'{e}}maux and Wulfram Gerstner.
    \newblock {Neuromodulated spike-timing-dependent plasticity, and theory of
      three-factor learning rules}.
    \newblock {\em Frontiers in Neural Circuits}, 9:85, 2016.
    
    \bibitem{lichtsteiner2008128}
    Patrick Lichtsteiner, Christoph Posch, and Tobi Delbruck.
    \newblock {A 128 $\times$ 128 120 dB 15$\mu$ s latency asynchronous temporal
      contrast vision sensor}.
    \newblock {\em IEEE Journal of Solid-State Circuits}, 43(2):566--576, 2008.
    
    \bibitem{sironi2018hats}
    Amos Sironi, Manuele Brambilla, Nicolas Bourdis, Xavier Lagorce, and Ryad
      Benosman.
    \newblock {HATS: Histograms of averaged time surfaces for robust event-based
      object classification}.
    \newblock In {\em Proceedings of the IEEE Conference on Computer Vision and
      Pattern Recognition}, pages 1731--1740, 2018.
    
    \bibitem{lagorce2016hots}
    Xavier Lagorce, Garrick Orchard, Francesco Galluppi, Bertram~E Shi, and Ryad~B
      Benosman.
    \newblock {Hots: a hierarchy of event-based time-surfaces for pattern
      recognition}.
    \newblock {\em IEEE Transactions on Pattern Analysis and Machine Intelligence},
      39(7):1346--1359, 2016.
    
    \bibitem{afshar2020event}
    Saeed Afshar, Nicholas Ralph, Ying Xu, Jonathan Tapson, Andr{\'{e}} van Schaik,
      and Gregory Cohen.
    \newblock {Event-based feature extraction using adaptive selection thresholds}.
    \newblock {\em Sensors}, 20(6):1600, 2020.
    
    \bibitem{tapson2013synthesis}
    Jonathan~C Tapson, Greg~Kevin Cohen, Saeed Afshar, Klaus~M Stiefel, Yossi
      Buskila, Tara~Julia Hamilton, and Andr{\'{e}} van Schaik.
    \newblock {Synthesis of neural networks for spatio-temporal spike pattern
      recognition and processing}.
    \newblock {\em Frontiers in Neuroscience}, 7:153, 2013.
    
    \bibitem{Stein1967}
    R.~B. Stein.
    \newblock {Some Models of Neuronal Variability}.
    \newblock {\em Biophysical Journal}, 7(1):37--68, 1967.
    
    \bibitem{Stein1965}
    Richard~B. Stein.
    \newblock {A Theoretical Analysis of Neuronal Variability}.
    \newblock {\em Biophysical Journal}, 5(2):173--194, 1965.
    
    \bibitem{fourcaud2003spike}
    Nicolas Fourcaud-Trocm{\'{e}}, David Hansel, Carl {Van Vreeswijk}, and Nicolas
      Brunel.
    \newblock {How spike generation mechanisms determine the neuronal response to
      fluctuating inputs}.
    \newblock {\em Journal of Neuroscience}, 23(37):11628--11640, 2003.
    
    \bibitem{gerstner2002spiking}
    Wulfram Gerstner and Werner~M Kistler.
    \newblock {\em {Spiking neuron models: Single neurons, populations,
      plasticity}}.
    \newblock Cambridge university press, 2002.
    
    \bibitem{izhikevich2004model}
    Eugene~M Izhikevich.
    \newblock {Which model to use for cortical spiking neurons?}
    \newblock {\em IEEE Transactions on Neural Networks}, 15(5):1063--1070, 2004.
    
    \bibitem{kasabov2010spike}
    Nikola Kasabov.
    \newblock {To spike or not to spike: A probabilistic spiking neuron model}.
    \newblock {\em Neural Networks}, 23(1):16--19, 2010.
    
    \bibitem{jolivet2004generalized}
    Renaud Jolivet, Timothy~J Lewis, and Wulfram Gerstner.
    \newblock {Generalized integrate-and-fire models of neuronal activity
      approximate spike trains of a detailed model to a high degree of accuracy}.
    \newblock {\em Journal of Neurophysiology}, 92(2):959--976, 2004.
    
    \bibitem{hodgkin1952quantitative}
    Alan~L Hodgkin and Andrew~F Huxley.
    \newblock {A quantitative description of membrane current and its application
      to conduction and excitation in nerve}.
    \newblock {\em The Journal of Physiology}, 117(4):500--544, 1952.
    
    \bibitem{jolivet2006predicting}
    Renaud Jolivet, Alexander Rauch, Hans-Rudolf L{\"{u}}scher, and Wulfram
      Gerstner.
    \newblock {Predicting spike timing of neocortical pyramidal neurons by simple
      threshold models}.
    \newblock {\em Journal of Computational Neuroscience}, 21(1):35--49, 2006.
    
    \bibitem{bohte2002unsupervised}
    Sander~M Bohte, Han {La Poutr{\'{e}}}, and Joost~N Kok.
    \newblock {Unsupervised clustering with spiking neurons by sparse temporal
      coding and multilayer RBF networks}.
    \newblock {\em IEEE Transactions on Neural Networks}, 13(2):426--435, 2002.
    
    \bibitem{wade2010swat}
    John~J Wade, Liam~J McDaid, Jose~A Santos, and Heather~M Sayers.
    \newblock {SWAT: a spiking neural network training algorithm for classification
      problems}.
    \newblock {\em IEEE Transactions on Neural Networks}, 21(11):1817--1830, 2010.
    
    \bibitem{ghosh2007improved}
    Samanwoy Ghosh-Dastidar and Hojjat Adeli.
    \newblock {Improved spiking neural networks for EEG classification and epilepsy
      and seizure detection}.
    \newblock {\em Integrated Computer-Aided Engineering}, 14(3):187--212, 2007.
    
    \bibitem{gueorguieva2006learning}
    Natacha Gueorguieva, Iren Valova, and Georgi Georgiev.
    \newblock {Learning and data clustering with an RBF-based spiking neuron
      network}.
    \newblock {\em Journal of Experimental \& Theoretical Artificial Intelligence},
      18(01):73--86, 2006.
    
    \bibitem{yu2014brain}
    Qiang Yu, Huajin Tang, Kay~Chen Tan, and Haoyong Yu.
    \newblock {A brain-inspired spiking neural network model with temporal encoding
      and learning}.
    \newblock {\em Neurocomputing}, 138:3--13, 2014.
    
    \bibitem{lecun1998gradient}
    Yann LeCun, L{\'e}on Bottou, Yoshua Bengio, and Patrick Haffner.
    \newblock Gradient-based learning applied to document recognition.
    \newblock {\em Proceedings of the IEEE}, 86(11):2278--2324, 1998.
    
    \bibitem{diehl2015unsupervised}
    Peter~U Diehl and Matthew Cook.
    \newblock {Unsupervised learning of digit recognition using
      spike-timing-dependent plasticity}.
    \newblock {\em Frontiers in Computational Neuroscience}, 9:99, 2015.
    
    \bibitem{merolla2011digital}
    Paul Merolla, John Arthur, Filipp Akopyan, Nabil Imam, Rajit Manohar, and
      Dharmendra~S Modha.
    \newblock A digital neurosynaptic core using embedded crossbar memory with 45pj
      per spike in 45nm.
    \newblock In {\em 2011 IEEE custom integrated circuits conference (CICC)},
      pages 1--4. IEEE, 2011.
    
    \bibitem{neftci2014event}
    Emre Neftci, Srinjoy Das, Bruno Pedroni, Kenneth Kreutz-Delgado, and Gert
      Cauwenberghs.
    \newblock Event-driven contrastive divergence for spiking neuromorphic systems.
    \newblock {\em Frontiers in neuroscience}, 7:272, 2014.
    
    \bibitem{hussain2014improved}
    Shaista Hussain, Shih-Chii Liu, and Arindam Basu.
    \newblock Improved margin multi-class classification using dendritic neurons
      with morphological learning.
    \newblock In {\em 2014 IEEE International Symposium on Circuits and Systems
      (ISCAS)}, pages 2640--2643. IEEE, 2014.
    
    \bibitem{beyeler2013categorization}
    Michael Beyeler, Nikil~D Dutt, and Jeffrey~L Krichmar.
    \newblock Categorization and decision-making in a neurobiologically plausible
      spiking network using a stdp-like learning rule.
    \newblock {\em Neural Networks}, 48:109--124, 2013.
    
    \bibitem{zhao2014feedforward}
    Bo~Zhao, Ruoxi Ding, Shoushun Chen, Bernabe Linares-Barranco, and Huajin Tang.
    \newblock Feedforward categorization on aer motion events using cortex-like
      features in a spiking neural network.
    \newblock {\em IEEE transactions on neural networks and learning systems},
      26(9):1963--1978, 2014.
    
    \bibitem{tavanaei2019bp}
    Amirhossein Tavanaei and Anthony Maida.
    \newblock Bp-stdp: Approximating backpropagation using spike timing dependent
      plasticity.
    \newblock {\em Neurocomputing}, 330:39--47, 2019.
    
    \bibitem{lee2016training}
    Jun~Haeng Lee, Tobi Delbruck, and Michael Pfeiffer.
    \newblock Training deep spiking neural networks using backpropagation.
    \newblock {\em Frontiers in neuroscience}, 10:508, 2016.
    
    \bibitem{mostafa2017supervised}
    Hesham Mostafa.
    \newblock Supervised learning based on temporal coding in spiking neural
      networks.
    \newblock {\em IEEE transactions on neural networks and learning systems},
      29(7):3227--3235, 2017.
    
    \bibitem{iyer2021neuromorphic}
    Laxmi~R Iyer, Yansong Chua, and Haizhou Li.
    \newblock Is neuromorphic mnist neuromorphic? analyzing the discriminative
      power of neuromorphic datasets in the time domain.
    \newblock {\em Frontiers in neuroscience}, 15:297, 2021.
    
    \bibitem{tidigits}
    R~Leonard.
    \newblock A database for speaker-independent digit recognition.
    \newblock In {\em IEEE International Conference on Acoustics, Speech, and
      Signal Processing}, volume~9, pages 328--331. IEEE, 1984.
    
    \bibitem{Zhang_Wu_Chua_Luo_Pan_Liu_Li_2019}
    Malu Zhang, Jibin Wu, Yansong Chua, Xiaoling Luo, Zihan Pan, Dan Liu, and
      Haizhou Li.
    \newblock {MPD-AL: An Efficient Membrane Potential Driven Aggregate-Label
      Learning Algorithm for Spiking Neurons}.
    \newblock {\em Proceedings of the AAAI Conference on Artificial Intelligence},
      33(01):1327--1334, 2019.
    
    \bibitem{gutig2009time}
    Robert G{\"{u}}tig and Haim Sompolinsky.
    \newblock {Time-warp--invariant neuronal processing}.
    \newblock {\em PLoS Biology}, 7(7), 2009.
    
    \bibitem{shaban2021adaptive}
    Ahmed Shaban, Sai~Sukruth Bezugam, and Manan Suri.
    \newblock {An adaptive threshold neuron for recurrent spiking neural networks
      with nanodevice hardware implementation}.
    \newblock {\em Nature Communications}, 12(1):1--11, 2021.
    
    \bibitem{zhang2003other}
    Wei Zhang and David~J Linden.
    \newblock {The other side of the engram: experience-driven changes in neuronal
      intrinsic excitability}.
    \newblock {\em Nature Reviews Neuroscience}, 4(11):895, 2003.
    
    \end{thebibliography}

\end{document}